\documentclass[10pt,journal,compsoc]{IEEEtran}
\usepackage[ruled]{algorithm2e} 

\usepackage{amsfonts}
\usepackage{xcolor}
\usepackage{soul,color}
\usepackage[pdftex]{graphicx}

\usepackage{amsmath}
\usepackage{multirow}
\usepackage{pdfpages}

\ifCLASSOPTIONcompsoc
  \usepackage[nocompress]{cite}
\else
  \usepackage{cite}
\fi
\begin{document}

%
\title{Sketch2Stress: Sketching with Structural Stress Awareness}
%
%
%
%

\author{Deng~Yu,
        Chufeng~Xiao,
        Manfred~Lau\IEEEauthorrefmark{1},
        and~Hongbo~Fu\thanks{\IEEEauthorrefmark{1} Corresponding authors}\IEEEauthorrefmark{1}
\IEEEcompsocitemizethanks{
\IEEEcompsocthanksitem Deng Yu, Chufeng Xiao, Manfred Lau, and Hongbo Fu are with the School
of Creative Media, City University of Hong Kong.\protect\\
E-mail: \{deng.yu, chufeng.xiao\}@my.cityu.edu.hk, \{manfred.lau, hongbofu\}@cityu.edu.hk}
\thanks{Manuscript received xx xx, xx; revised xx xx, xx.}}

%
%

\markboth{IEEE TRANSACTIONS ON VISUALIZATION AND COMPUTER GRAPHICS}%
{\MakeLowercase{\textit{et al.}}: Bare Demo of IEEEtran.cls for Computer Society Journals}
%



\IEEEtitleabstractindextext{%
\begin{abstract}
In the process of product design and digital fabrication, 
{the} structural analysis
{of a}
designed prototype
is a fundamental and essential step.
However, such {a} step is usually invisible or inaccessible
to designers
at the early sketching phase.
This limits the user's ability to consider a shape's physical properties and structural soundness.
To bridge this gap, we introduce a novel approach \textit{Sketch2Stress} that allows users to perform structural analysis of {desired} 
objects at the sketching stage.
This method takes as input a {2D freehand} sketch and {one or multiple locations of user-assigned external forces.}
{With the specially-designed two-branch generative-adversarial framework, it} automatically predicts a normal map and a corresponding structural stress map distributed over the user-sketched underlying object.
In this way, our method empowers designers to easily examine the stress sustained everywhere and identify 
potential problematic regions {of} their sketched object.
Furthermore, combined with the predicted normal map, users are able to conduct a region-wise structural analysis efficiently by aggregating the stress effects of multiple forces in the same {direction}. 
{Finally,} we demonstrate the effectiveness and practicality of our system with extensive experiments and 
user studies.
\end{abstract}

}

\maketitle

\IEEEdisplaynontitleabstractindextext

%
\IEEEpeerreviewmaketitle

\IEEEraisesectionheading{
}
\section{Introduction}
\label{section:intro}

\begin{figure*}[t]{
    \centering
  \includegraphics[width=.96\textwidth]{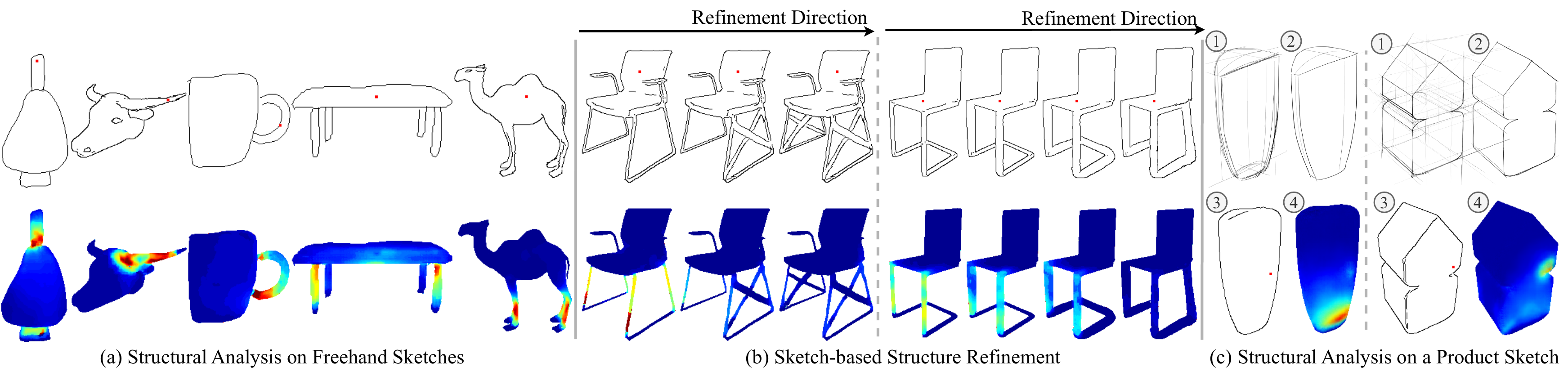}
  \caption{Our \emph{Sketch2Stress} system supports users to easily perform structural analysis  on their freely sketched objects {by assigning forces at desired locations} {(shown in red dots)} (a), and structural refinement ({in each example}, the upper {row shows} 
  the {progressively} refined sketches while the bottom {row shows} 
  our computed stress maps) on the weak regions of 
  problematic sketched object{s} {with real-time feedback of a stress map along with their editing operations} (b). We also show that our system can handle professional product sketches, e.g. those in the OpenSketch dataset (c){, after a separate training process.}
  {In} (c),  we illustrate {two}
  example{s} of using professional product sketches for structural analysis, starting from 
  {the}  concept {sketches}, then 
  {the} presentation {sketches}, 
  {the}   clean {sketches}, 
  and finally, our generated structural stress {maps} {under the applied forces (red dots)}.
  }
  \label{fig:teaser}
  }
\end{figure*}

\IEEEPARstart{T}{he} design and fabrication process typically begins with sketching on paper, followed by
{digitization}, 
and eventually, the use of fabrication machineries such as a waterjet, laser cutter, or 3D printer \cite{johnson2012sketch}.
Conventional structural analysis is used in both the digitization and manufacturing stages in a trial-and-error manner{. This is a costly process}, 
in terms of time, labor, and materials.
To facilitate product design and digital fabrication, numerous structural analysis techniques \cite{stava2012stress,zhou2013worst,wang2013cost, prevost2013make, ulu2017lightweight, panetta2017worst, lu2014build} {have been} 
proposed to simulate the physical environment and directly analyze or optimize digital prototype structure{s} virtually at the digitization stage.
The goals of these techniques can be generally categorized into several aspects: weakness analysis \cite{zhou2013worst,panetta2017worst,langlois2016stochastic}, structural enhancement \cite{stava2012stress,miki2015parametric}, inner or surface material optimization \cite{ulu2017lightweight, prevost2013make, lu2014build, zehnder2016designing, dumas2015example, fang2020reinforced}, and specified
properties \cite{yao2015level, prevost2013make, schumacher2015microstructures}.
While these structural analysis tools
enrich product design and fabrication, they are less accessible to 
designers at the early sketching stage since the effect of external physical factors on {an object being designed} 
is unknown to users during sketching.

In this work, we study the structural analysis of {a sketched object {and use the resulting analysis}} 
to generate the stress effect of the object under external forces at specified locations{, as displayed in Figure \ref{fig:teaser}}.
Addressing this problem could enable designers to notice the potential structural weakness, specify their design space under different force configurations, and further refine {the object} 
at the sketching stage.
Furthermore, this will open up possibilities for promoting sketch-based design and diagnosis to non-experts since
sketching is an intuitive and universal tool for creativity and expression for novice users.

Since the existing digital structural analysis methods are mainly performed on 3D prototypes, a straightforward strategy to solve our problem 
might be 
to first use sketch-based shape reconstruction methods \cite{lun20173d,li2018robust,delanoy20183d, smirnov2021learning, zhang2021sketch2model, guillard2021sketch2mesh, wang2018pixel2mesh},  
followed by a 3D structural analysis method.
However, 
existing sketch-based shape reconstruction approaches
{suffer several common limitations.}
{First, they require} 
specified
multi-view sketches of the same object as input \cite{lun20173d,li2018robust}, but their creation is 
highly demanding for users{. Second, when taking a single-view sketch as input, they often demand} 
additional conditions such as camera parameters \cite{zhang2021sketch2model,guillard2021sketch2mesh} or 3D deformable templates \cite{smirnov2021learning, wang2018pixel2mesh}{, making it difficult to reconstruct}  
shapes with complex structures from one 
sketch only.
Since 3D shape reconstruction is a difficult task, we are interested in directly performing  
structural analysis 
based on only an input sketch and the external force conditions.









While performing a structural analysis method \cite{ulu2017lightweight} on 3D shapes, we observed that:
(i) shapes with similar structures have similar stress distributions under the same external force with the same location, direction, and magnitude; and
(ii) on the same shape, neighboring points in local regions undertake {similar} 
stress under an external force.
This makes it possible to use a data-driven strategy to solve our problem.
Therefore, we further transform the problem of sketch-based structural analysis into {an} 
image-to-image translation problem \cite{isola2017image, wang2018high}, where we leverage a neural network to learn the mapping from 
input sketches to structural analysis results conditioned on external forces.

Since there is no existing dataset for sketch-based structural analysis,
we construct a novel {sketch-force-stress}
dataset by first defining rules to normalize and uniform 
force regions on 3D shapes in the same category based on {Observation} 
(i).
Based on Observation (ii), 
{we then} {uniformly sample 200 $\sim$ 300 key force locations {(sampling more force locations helps increase the accuracy of structural stress map computation for more detailed geometry 
but would incur heavier computation burdens)} on the surface of 3D shapes in each view to analyze their structural soundness rather than exhaustively sampling all the surface points,}
and apply 3D structural analysis to the
shapes{,} {where the external forces are set with equivalent magnitude \cite{ulu2017lightweight}}, and {finally} 
render 
multi-view sketches and the {corresponding} view-dependent stress maps 
from the 3D structural analysis results.
In this way, we collect a large-scale dataset consisting of quadruples of an input sketch, a point map indicating {the} force location, 
a normal map recording force directions {of all possible forces}, 
and a corresponding structural stress map. 
Note that {inheriting the assumption of \cite{ulu2017lightweight},} we 
set
the \textit{magnitude} of external forces {in our problem} to be all the same 
{and set the force \textit{directions} to be the same as the surface normals (pointing inward) at the force locations}.
{Also, {since} 
Ulu et al. \cite{ulu2017lightweight} {rely} 
on the boundary shell to represent the shape structures of 3D models and further approximate the relationship between input forces and resulting stresses on this representation, 
the same boundary shell representation is 
inherited implicitly in our assumption for {user-designed} 
objects.
Therefore, {sketched objects corresponding to commonly seen real-world objects might exhibit} 
severely fragile regions {({see the "problematic structure" in} Figure {\ref{fig:user_study_with_tool}}) since} 
the inner material properties and inter-part connection manners are not considered in the boundary shell {setting}. 
}





To synthesize a structural stress map from an input sketch conditioned on an external force with arbitrary \textit{{location}} and uncertain \textit{{direction}}, we present a novel framework combining a one-encoder-{two}-branch-decoder generator {with} 
two discriminators: one branch in the generator is used to synthesize a 
corresponding 
structural stress map from the input sketch and the 
force location; the other branch aims to infer the direction {(opposite-normal)} of the external force. 
These two branches jointly guarantee that the generator can perceive the distinctive location{s} and direction{s} of external forces imposed on sketches.
Two discriminators supervise the learning process of {the} two branches of the generator.

{With our trained {network}, users can easily check the structural soundness of a sketched object under {a} 
single force assigned at any location.} 
In addition, a well-known physical axiom states that: "If two forces act on an object in the same direction, the net force is equal to the sum of the two forces".
Based on this axiom, we present an efficient region-wise sketch-based structural analysis method to approximate the stress effect of the net force at a local region by aggregating the stress maps of multiple forces {at different locations but} in the same normal direction.

Our contributions can be summarized as follows:
\begin{itemize}
\item To the best of our knowledge, we are the first to study the problem of sketch-based {shape} structural analysis.
\item We introduce a novel {two}-branch 
generator to learn the mapping from {a user-drawn} 
sketch conditioned on the external force variables to {a} 
structural stress map.
\item We present a sketch-based structural analysis interface that supports structural weakness detection and structural refinement on sketches.
\item We collect a large-scale sketch-based structural analysis dataset containing millions of {sketch-force-stress data triplets}
spanning 11 shape categories. We will release the data and code to the research community.
\end{itemize}

\begin{figure}[t]{
    \centering
    \includegraphics[width=.96\linewidth]{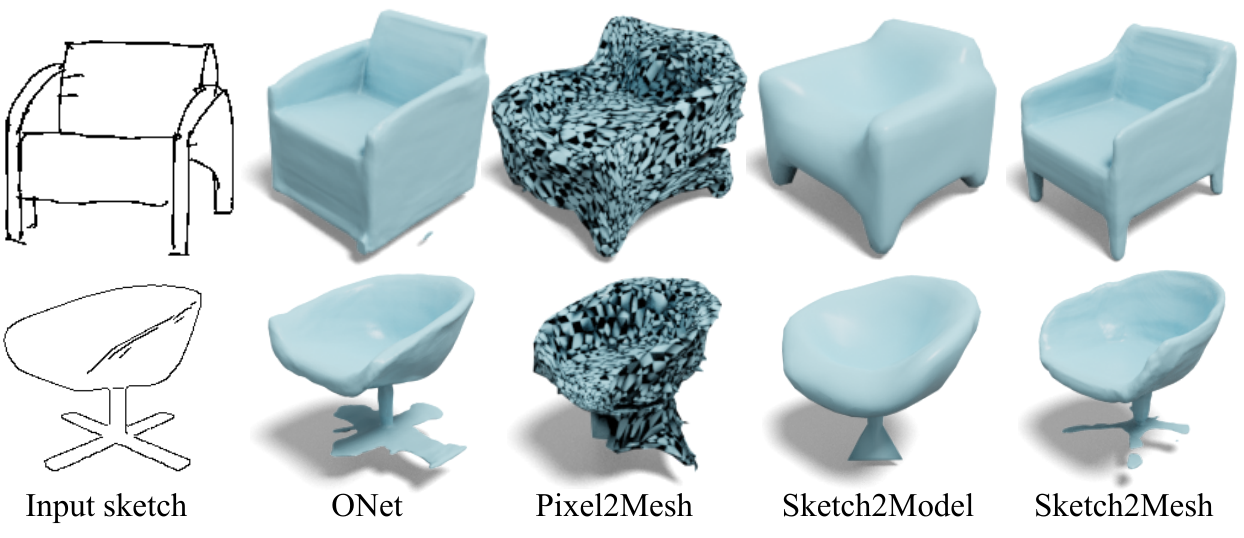}
    \caption{Single-view sketch-based shape reconstruction methods. {We can see ONet \cite{mescheder2019occupancy}, Pixel2Mesh, Sketch2Model, and Sketch2Mesh} 
    fail to reconstruct the geometry details of the input sketches. 
    {ONet and Sketch2Mesh tend to generate detached noises, broken parts, and the inconsistent orientation of chair legs at the bottom row. While Pixel2Mesh and  Sketch2Model generate too coarse shape results where {the former's} 
    surface patches are widely corrupted, and the {latter's} 
    local details are heavily over-smoothed.}
}
    \label{fig:sk2mesh}
    }
\end{figure}

\section{Related Work}
\label{section:related_work}
{In this section, we review the previous works closely related to us, namely, 3D structure analysis, sketch-based shape reconstruction, and image-to-image translation.}

\subsection{{3D Structure Analysis}}
{Various} 
recent works support computational analysis on the structural soundness of 3D shapes.
Especially with the emergence of 3D printing techniques, numerous approaches 
were proposed for 
printed objects 
in a wide range of tasks, from structural weakness detection \cite{ulu2017lightweight, stava2012stress, zhou2013worst, panetta2017worst} to material optimization \cite{prevost2013make, lu2014build, zehnder2016designing, dumas2015example, fang2020reinforced}.
Since {it is} 
challenging to convey the variations of materials {from one sketch}, 
we do not review the material-oriented approaches.

The first structural analysis work for {3D} printed 
objects dates back to \cite{telea2011voxel}, where {Telea and Jalba} 
identify thin and thick parts and {estimate} 
whether a thin part could support its attached parts under {several} 
pre-defined geometric rules. 
Later, Stava et al. \cite{stava2012stress} use FEM (finite element method \cite{hughes1987finite}) to discover and strengthen problematic components of {a} 
printed model under the applied gravity load and 2-finger gripping loads. 
Then Zhou et al. \cite{zhou2013worst} propose an analysis technique to predict 
fragile regions under worst-case external force loads by identifying 
potential regions of a structure that might fail under arbitrary force configurations. 
Later, Langlois et al. \cite{langlois2016stochastic} present a stochastic {FEM} 
{for} predicting failure probabilities under the force loads at contact regions.
Different from previous works with specified force {load} 
setting{s}, Ulu et al. \cite{ulu2017lightweight} propose a more general structural optimization approach that examines 3D shapes with any force loads at arbitrary locations and computes a feasible material distribution to withstand such forces.
Different from 3D prototypes, {sketches} are 
usually created in the 2D space with sparse content. 
This makes it difficult to apply 
finite element analysis, which is the basic technique for most 3D structural analysis approaches. 
{To model the relationship between the input forces and the corresponding structural stress of the sketched objects, we translate the sketch-based structure analysis problem to the data-driven image-to-image translation task where we learn the mapping between the input sketches and the structural stress responses conditioned on the variable external forces from massive 
{sketch-force-stress data triplets}. 
}

\subsection{{Sketch-based Shape Reconstruction}}
Using an additional step to convert input sketches to intermediate 3D shapes with sketch-based shape reconstruction approaches usually requires extra conditions such as multi-view inputs, camera parameters, and 3D category templates, which are lacking in our scenario.
{Figure \ref{fig:sk2mesh} 
further displays the limitations of state-of-the-art single-view sketch-based shape reconstruction methods. The 3D meshes generated by ONet 
\cite{mescheder2019occupancy} and Pixel2Mesh \cite{wang2018pixel2mesh} have obvious artifacts, like detached parts and inverse patches, which prevent performing FEM requiring continuous and closed input 3D surfaces. 
The meshes generated by Sketch2Model \cite{zhang2021sketch2model} lose too many geometry details.
For Sketch2Mesh \cite{guillard2021sketch2mesh}, its {bottom} 
reconstructed mesh has not only broken parts but also inconsistent legs compared with the input sketch. 
Hence, none of {these} 
approaches could {perform shape reconstruction from sketches robustly}}. 
Therefore, we turn to image-to-image translation techniques
to directly generate a feasible 2D structural analysis result for an input sketch. 
{To demonstrate the faithfulness and effectiveness of our sketch-based structure analysis approach, we provide a direct comparison between our Sketch2Stress approach and the reconstruction-and-simulation approach applying stress simulation \cite{ulu2017lightweight} on generated meshes (Figure \ref{fig:sk2mesh}) {by} 
{ONet}/Sketch2Model/Sketch2Mesh after post-cleanup.
Compared with the reconstruction-and-simulation results, {as shown in Figure \ref{fig:recons_and_simulate}}, our method can reconstruct a view-dependent structure robustly 
{and} is more competent for the sketch-based structure analysis task than the reconstruction-and-simulation way.

\begin{figure}[t]{
    \centering
    \includegraphics[width=1.01\linewidth]{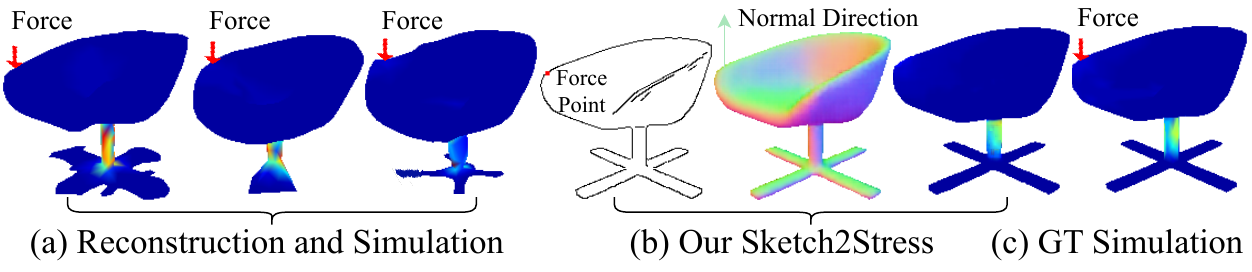}
    \caption{
    {Comparison of the reconstruction-and-simulation  way (a) and our Sketch2Stress (b). {The models in (a) are the reconstructed meshes in Fig. \ref{fig:sk2mesh} (Bottom).}
    The red arrows {indicate} 
    the applied external forces. In (b), the force is plotted on the input sketch, and the generated normal map and stress map are side-placed. The ground-truth 3D stress simulation {is given} 
    in (c). Please zoom in to examine the details of the above stress distributions.}
    }
    \label{fig:recons_and_simulate}
    }
\end{figure}

\subsection{{Image-to-Image Translation}}
Since Isola et al. \cite{isola2017image} and Wang et al. \cite{wang2018high} introduced the general-purpose cGAN frameworks for diverse types of inputs, e.g., realistic images, sketch{es}, and semantic masks,
there are many sketch-based image synthesis tasks using image-to-image translation techniques.
The most related {to ours}
are 3D-aware approaches \cite{su2018interactive, jiao2020tactile} with sketch inputs.  
For instance, Su et al. \cite{su2018interactive} present an interactive system for high-quality normal map generation. 
Later, Jiao et al. \cite{jiao2020tactile} propose a joint framework that leverages category and depth information to improve shape understanding for 
tactile sketch saliency prediction. 
However, the aforementioned methods cannot be directly applied to our problem since they have no proper way to represent the external forces with their designed frameworks.

\begin{figure*}[t]{
    \centering
    \includegraphics[width=.9\linewidth]{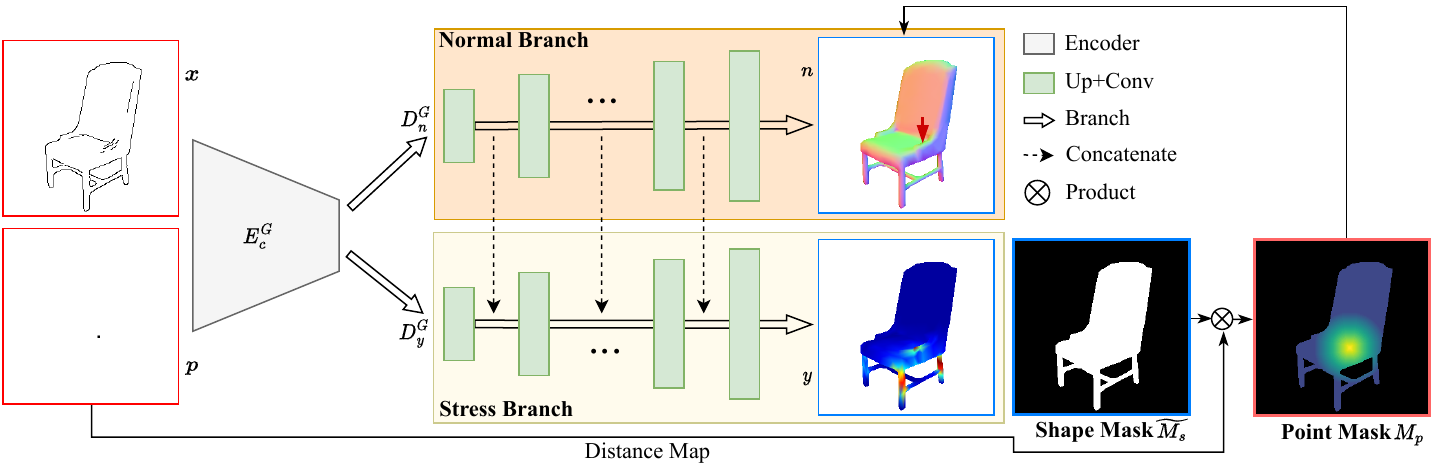}
    \caption{Overview of the 
    {two}-branch generator of Sketch2Stress. 
             Given {an} 
             input 
             sketch (upper left) and an input point map (lower left) indicating {a force location}, 
             the {two}-branch generator uses its encoder to learn a sketch-force joint feature space, {and} then leverage{s} 
             {two} decoders to synthesize the corresponding stress map (lower branch) {and a normal map (upper branch). We use warmer colors (reds and yellows) to show high stress and cooler colors (greens and blues) to show low stress. {Our framework has four distinctive functional modules or elements {highlighted in boldface}: 
             Normal Branch, Stress Branch, Shape Mask, and Point Mask. 
             The Normal Branch infers not only the force direction (in the opposite-normal direction{, illustrated by a} 
             red arrow) at the input force map $p$ but also an entire view-dependent normal map $n$ indicating the underlying geometry 
             from the input sketch $x$. As for the role of the Stress Branch, it gathers information of the sketched object and how the force is  applied from  the sketch-force feature space and the shared Normal Branch's feature space, producing the final  stress map $y$.} 
             {{The} 
             predicted shape mask $\widetilde{M_s}$ and the point-attention mask $M_p$ }are proposed to further {regularize} the shape boundaries and {enhance the spatial information of} {the user-specified} force {location} during the generation process. }}
    \label{fig:network}
    }
\end{figure*}
{Recently, diffusion models \cite{sohl2015deep,ho2020denoising,song2020score} have been used to obtain state-of-the-art results in {text-to-image synthesis and} text-guided image {editing}  \cite{meng2021sdedit, nichol2021glide, ramesh2022hierarchical}.
{The aforementioned diffusion models commonly rely on a Markov chain of diffusion steps to generate high-quality images from noises.}
{Despite} 
the impressive and realistic generation performance of the aforementioned methods, it is still challenging for diffusion models to impose precise spatial control on the generation outputs due to the nature of the one-to-one mapping between 
noise vectors 
and the corresponding ground-truth data samples. 
This limitation makes diffusion models unsuitable for our problem, which requires precise and faithful control over force locations.
Additionally, diffusion models generate images from noise vectors through 
iterative intermediate steps during the inference (denoising) stage, 
consuming more time than {these image-to-image} 
networks \cite{isola2017image, wang2018high}. 
Therefore, existing diffusion models might not be suitable for our real-time editing scenario.}

\section{Methodology}
\label{section:methodology}

In this work, we focus on the study of structural analysis {of sketched objects under external forces at user-specified locations}. 
{{Adapting} 
the structural analysis task from informative 3D objects to 2D sketches is nontrivial due to 
the ill-posed nature of sparse sketches {to represent continuous and closed 3D surfaces} 
as well as the challenge of representing {external forces} 
applied to the sketched objects.}
{To address these issues, we simplify the problem and make our assumptions as follows:
(i) We decouple the external forces to the constant force magnitude of $100 N$ 
and directions based on the estimation of a normal map (the force direction and the normal direction at the force location are opposite in our approach, as shown in Figure \ref{fig:recons_and_simulate} (b)). 
(ii) Then we utilize an effective {data-driven} way to approximate the mathematically/physically precise stress by constructing a novel large-scale sketch-force-stress dataset and proposing a new 
{two}-branch (for force location and direction) generation pipeline (see Figure \ref{fig:network}).}
{(iii)} Note that the materials of the
sketched objects are assumed to be the same with linear isotropic materials and small deformations, 
following \cite{ulu2017lightweight}.

To faithfully represent the external forces applied to sketched objects, we utilize a set of 2D point maps $P$
to specify the force locations {(one point map for each force location)} 
and a 
2D normal map 
$n \in {\mathbb{R}}^{256 \times 256 \times 3}$ {of each view of an object} to record the force {direction} $-n_p$ at the corresponding location $p$. 
In this way, we decouple the original 3D external forces into the above 2D representations  
that can be further treated as conditions for {mapping} 
the input sketches $X$
to the corresponding stress maps $Y$.
{Let} {$N$ {denote} 
the set  
{of normal maps.}

{As illustrated in Figure \ref{fig:network},} 
Our framework for sketch-based structural analysis consists of two components: (1) a
{two}-branch 
generator $G$$:$ $(x, p) \rightarrow (n, y)$, 
including a common sketch-force encoder $E_c^G$ and two separated decoders $D_n^G$ and $D_y^G$ for {the} normal map $n$ and {the} stress map $y$, illustrated in Figure \ref{fig:network}, 
and 
(2) two multi-scale discriminators $D_n$  and $D_y$ for normal and stress maps.
Specifically, given an input sketch and the condition of a point map, 
the common encoder of our 
{two}-branch generator constructs a joint feature space $E_c^G(x, p)$ for the input sketch and the 
{input point map}.
The {subsequent}
two decoders (7-layer up-sampling and convolution) $D_n^G$ and $D_y^G$  infer the correct normal directions $\widetilde{n}$ and a feasible structural stress map $\widetilde{y}$ from this joint feature space, respectively.
These two branches enforce that the common encoder $E_c^G$ should learn a joint feature representation that captures not only the geometry and normal directions of the input sketch but also the distinctive force location on the input point map. 
Note that {the} feature maps in the normal decoder $D_n^G$ are layer-wise concatenated to the stress decoder $D_y^G$ to enrich its structure perception.  
Finally, {the} two multi-scale discriminators
distinguish real images from the translated ones at $256 \times 256$, $128 \times 128$, and $64 \times 64$ scales. 
This is a standard way to represent distinctive, fine-grained details in images \cite{wang2018high}. 
Finally, we jointly optimize G, $D_n$, and $D_y$ with the {following GAN loss}: 
\begin{equation}
 \begin{split}
L_{G, D} & = \mathbb{E}_y[\log D_y(y)] + \mathbb{E}_n[\log D_n(n)] \\ & \quad +  \mathbb{E}_{x,p}[\log(1-D_n(G(x,p))) + \log(1-D_y(G(x,p))]. 
\end{split}
\label{eqn:GAN}
\end{equation}
where $x$, $p$, $n$, and $y$ 
refer to the quadruple of an input sketch, a point map, a normal map, and the corresponding stress map.

\textit{{\textbf{Shape Constraints}.}}
To overcome the issue that the generated pixels are often outside of the shape boundary of the sketched objects in the stress maps, we further predict a one-channel shape mask $\widetilde{M_s}$ {(Figure \ref{fig:network})} from the joint feature space $E_c^G(x, p)$. This shape mask is also useful for reducing shape ambiguity in normal {map generation}. 
{Therefore, we use a shape loss $L_{shape}$ to measure the $L1$ distance between {a} generated shape {mask} 
and the ground truth, as {formulated} 
below:}
\begin{equation}
L_{shape} = L1(\widetilde{M_s},{M_s}).
\label{eqn:shapeloss}
\end{equation}

\textit{\textbf{Force-point Constraints}.}
To emphasize the importance of {a force location}  
in the point map, we compute a point attention map $M_p$ by multiplying a point-centered distance map \footnote{{This is computed by
$D(Q)[q_f]=\frac{max_D-dist(q_i,q_f)}{max_D}, \ q_i\in Q$, where $q_i$ and $q_f$ are the locations of {every spatial point} 
and the force point of a point map{, respectively}. $Q$, $dist$, and $max_D$ are a set of all pixels in the point map, Euclidean distance, and the largest distance between $q_i$ and $q_f${, respectively}.
}} 
{(emphasiz{ing} the spatial importance of regions that surround the assigned force point)} with the shape mask.
We multiply this {attention map} 
{respectively} with the normal map and the stress map to {ensure} 
that the synthesized stress and normal directions surrounding this force point should be consistent with the ground-truth values as much as possible. 
{Here, we design a loss term $L_{point}$ to compute the $L1$ distance of the generated stress and normal maps compared with their {respective} ground truths inside the regions $M_p$, as defined below.} 
{
\begin{equation}
L_{point} = \lambda_1 L1(M_p \cdot \widetilde{y},\ M_p \cdot y) +  \lambda_2 L1(M_p \cdot \widetilde{n},\ M_p \cdot n). 
\label{eqn:pointloss}
\end{equation}
}
Therefore, our final objective function is formulated as follows:
{
\begin{equation}
L = L_{G,D} + \beta_1 L_{shape} + L_{point}, 
\label{eqn:totalloss}
\end{equation}
where we set $\lambda_1 = 100$, $\lambda_2 =100$, and $\beta_1 = 500$ in our experiments. The aforementioned setting achieves the best performance to balance the impact of different losses in our experiments.
Increasing $\lambda_1$, $\lambda_2$, and $\beta_1$ individually would force the network to partially focus on the different components, i.e., the quality of the generated normal map and stress map, or the predicted shape mask, respectively.}

\subsection{{Sketch-Force-Stress} Data Rendering}
\label{subsection:data_rendering}
To learn our network for sketch-based structural analysis, we need a considerably large dataset {of training data}. However, such a dataset is not available and expensive 
to acquire since it requires point-wise labeling for external forces and corresponding stress responses on the sketches.
Hence, we propose {to} {synthesize} 
the {sketch-force-stress data}
from existing 3D repositories, as shown in Figure \ref{fig:data_preparation}.

\begin{figure}[t]{
    \centering
    \includegraphics[width=.96\linewidth]{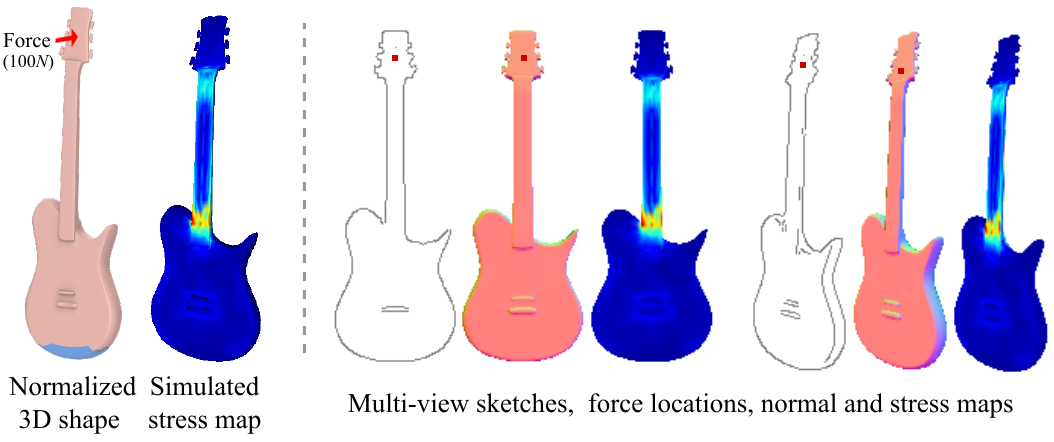}
    \caption{{Illustration for} data {preparation}.
    The left is a {normalized} 
    guitar {model} {(the bottom blue part is the fixed boundary condition and the upper is the contact regions)} and the 3D structure stress result under an external force {at a specific position}. 
    The right is the synthetic {sketch-force-stress}
    data.  We plot {the} force location on the 2D sketch{es} and normal map{s}.}
    \label{fig:data_preparation}
    }
\end{figure}
We first collect 
3D shapes from {several public shape repositories, including} ShapeNet \cite{chang2015shapenet}, AniHead \cite{du2020sanihead}, and COSEG \cite{wang2012active}. 
We convert 3D objects to watertight surfaces with \cite{stutz2020learning} {to make it ready for the {subsequent} 
3D structure analysis}. 
We then orient all 3D shapes {uprightly} {\cite{fu2008upright}}, 
move them onto the ground plane {(for fixing their bottom on the ground plane)}, and normalize them to a standard sphere.
{To normalize and uniform the force region{s} of different 3D shapes with diverse structures,}
{f}or shapes in the same category, we use the same ratio (0.02\% {$\sim$} 0.04\%) to define regions on 3D surfaces near the ground plane as fixed boundary conditions and the rest as contact regions allowing for any external forces (of $100N$ magnitude), 
as illustrated in Figure \ref{fig:data_preparation}.
Given {each 3D shape}, 
we 
uniformly sample force locations on {its} contact regions and {adopt the structural analysis approach in \cite{ulu2017lightweight} to} simulate the stress responses on {the shape's surface} 
under such forces in opposite-normal directions. 
Finally, we render the multi-view sketches, normal maps, force locations, and corresponding structural stress values $S$ from the simulated 3D stress results. 
{All of the above renderings are projected in the $256 \times 256$ spatial resolution.
The synthetic multi-view sketches are extracted from 2.5D normal maps using the Canny edge detector \cite{canny1986computational}. }
We project the 3D stress results with the azimuth angles of [$0, 45, 90$] degrees and the elevation angles in ($0\sim 15$) degrees.

Since the magnitude of the simulated stress value{s} 
$S$ spans an extremely large range from ten to ten million,
we further normalize the structural stress value{s} $S$ to a common $[0,1]$ space in two manners:
One is a shape-grained normalization to compare the fine-grained regional stress among single shapes; the other is a category-grained normalization to compare more general-grained shape stress among {all the} shapes in the same category.

\textit{\textbf{Shape-grained Normalization}.} 
{As mentioned before, similar shape structures tend to have similar weak regions under the same external forces. 
To highlight such region-wise stress similarity, we normalize the stress values inside each shape, as formulated below:}
{
\begin{equation}
        {S_i}' \leftarrow \frac{S_i}{\max(\{S_i\})},   \quad i \in a \ single \ shape. 
\label{eqn:stressdata}
\end{equation}}
where $i$ 
{is an index for} points in the contact regions of a 3D surface, and $S_i$ {is a stress value at the $i$-th point}. 

\textit{\textbf{Category-grained Normalization}.} 
{To study the general pattern (knowledge) of how different shape structures respond to the same external forces, we normalize the stress values of all the shapes in the entire dataset as Equation \ref{eqn:shapedata}.} 
{
\begin{equation}
     {S_j}'' \leftarrow \frac{{S_j}'}{\tau}, \quad   s.t. \ {S_j}'  \leftarrow \frac{S_j-u(S_j)}{\sigma(S_j)}, \ \ j  \in all \ shapes.  
\label{eqn:shapedata}
\end{equation}
}
where {$j$ {is the index} of points}  
in the contact regions on 3D surfaces, and $u(\cdot)$ and $\sigma(\cdot)$ are the mean and standard variance of the entire stress value set{, respectively}. $\tau =100$ is the upper boundary of the 99\% stress value. 
{Since $99\%$ stress values in ${S_j}'$ are smaller than $100$, we use $100$ as the upper boundary to filter out the stress with extremely high values, such as $1000$, $20000$, etc.} 
{Note that the resulting stress values in ${S_j}'$ have some negative values, indicating the extremely low-stress regions. Since we are 
interested {more} in the high-stress regions (fragile regions), we clip the positive value ($>0$) in ${S_j}'$ for further processing.}

\subsection{Region-wise Multi-force Aggregation}
\begin{figure}[t]{
    \centering
    \includegraphics[width=.96\linewidth]{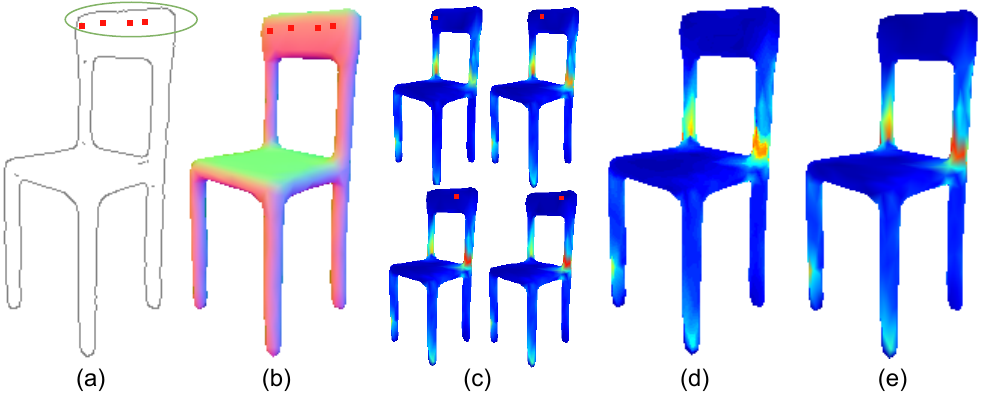}
    \caption{Pipeline of multi-force aggregation. (a) Input sketch and multi-force locations in a {local} 
    region. (b) Normal directions of multiple forces. (c) Four  
    stress maps {corresponding to each of the force locations}. (d) Aggregated stress effect from (c). (e) Ground-truth 3D simulation of multiple forces. 
    }
    \label{fig:multiforce}
    }
\end{figure}
With our trained network, users can easily {explore} 
the structural stress anywhere under a manually-assigned force {location} by clicking on the sketched object.
To further improve the efficiency of structural analysis on input sketches,  we provide a region-wise analysis method that aggregates the stress effects of multiple forces in a small 
region {along} 
the same normal directions.
After the user specifies a small region 
on a sketch (Figure \ref{fig:multiforce} (a)), with the predicted normal map, we automatically compute the 
force locations that have the same normal direction as the center point of this region (Figure \ref{fig:multiforce} (b)).
Then we directly add {and average} 
these stress effects (Figure \ref{fig:multiforce} (c)) together following the physical axiom in Section \ref{section:intro} and produce an aggregated stress map (Figure \ref{fig:multiforce} (d)). 
Compared with {a} 
3D simulated result (Figure \ref{fig:multiforce} (e)), although the overall stress effect in our aggregated stress map is diluted to some extent, it can still approximate the stress distribution of the 3D simulated result {well} and {can thus} be utilized as guidance for fragile detection.

\subsection{{Structural-Stress Awareness Replacement and Interpolation}}
{To demonstrate the sensitiveness of our Sketch2Stress to the variations in sketch structures, such as the significant structure changes (Figure \ref{fig:interpolation} (a)) and the more subtle geometry interpolations (Figure \ref{fig:interpolation} (b)), 
we first decompose an example chair into parts {and} then replace the original chair leg{s} with legs featuring significantly varied geometries and {by} linear{ly} interpolating the thickness values of the original chair leg{s}, respectively. 
Note that we keep the other parts of this example chair unchanged in these two tasks.
\begin{figure}[tb]{
    \centering
    \includegraphics[width=.96\linewidth]{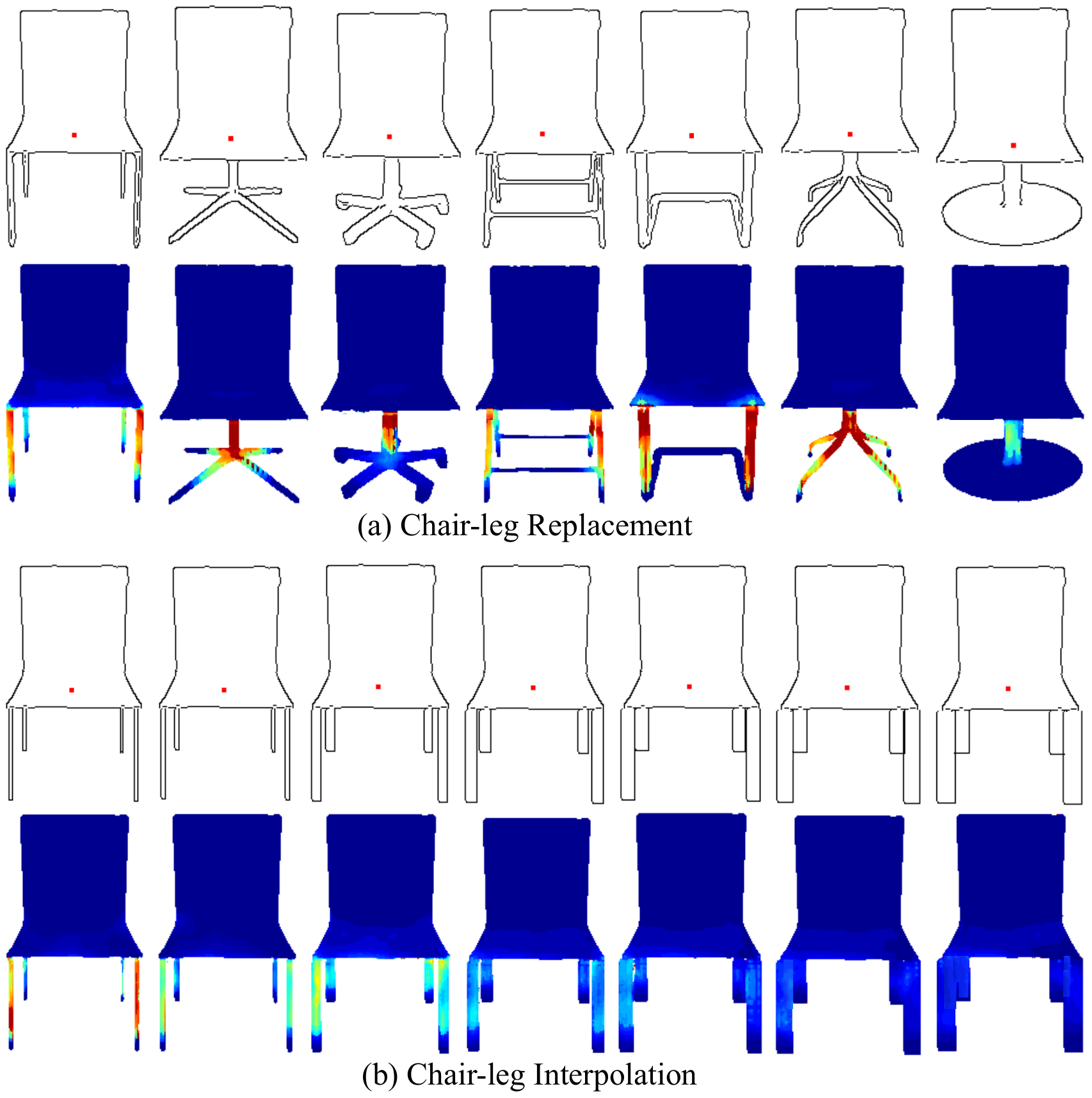}
    \caption{{Examples of our Sketch2Stress on Structure Replacement (a) and {Geometry} Interpolation (b).
    {Note that in (a) and (b), all the stress maps are in the same color range where the colors of the lower-stress regions are {closer} 
    to blue, while the higher-stress regions’ colors are {closer} 
    to red.}
    }}
    \label{fig:interpolation}
    }
\end{figure}As shown in Figure \ref{fig:interpolation} (a), the results are in line with our {expectation} 
that our well-trained Sketch2Stress is natural to perceive the structural soundness among highly changeable structures and identify corresponding fragile regions. 
Figure \ref{fig:interpolation} (b) further demonstrates our Sketch2Stress algorithm's capability in perceiving the tendency of thickness increment, distinguishing the subtle differences among highly similar structures, and generating the smooth stress distributions for those interpolated structures.
This could facilitate a sketch-based structural soundness suggestion task, where users could easily improve the structural soundness of their created sketches with our Sketch2Stress tool combined with the replacement and interpolation operations. 
}

\begin{figure}[tb]{
    \centering
    \includegraphics[width=.98\linewidth]{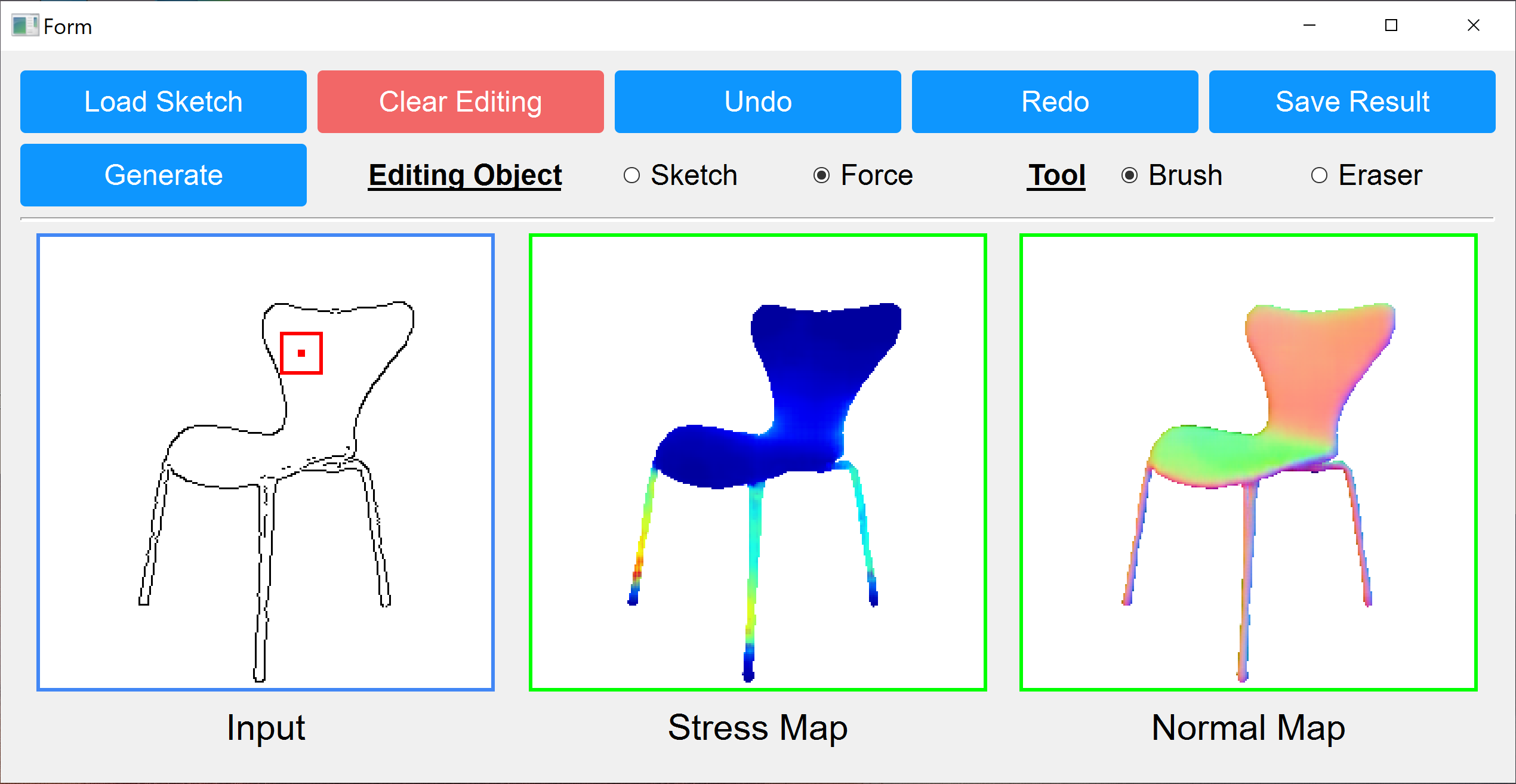}
    \caption{Our sketching interface.}
    \label{fig:interface}
    }
\end{figure}
\begin{figure*}[t]{
    \centering
    \includegraphics[width=.89\linewidth]{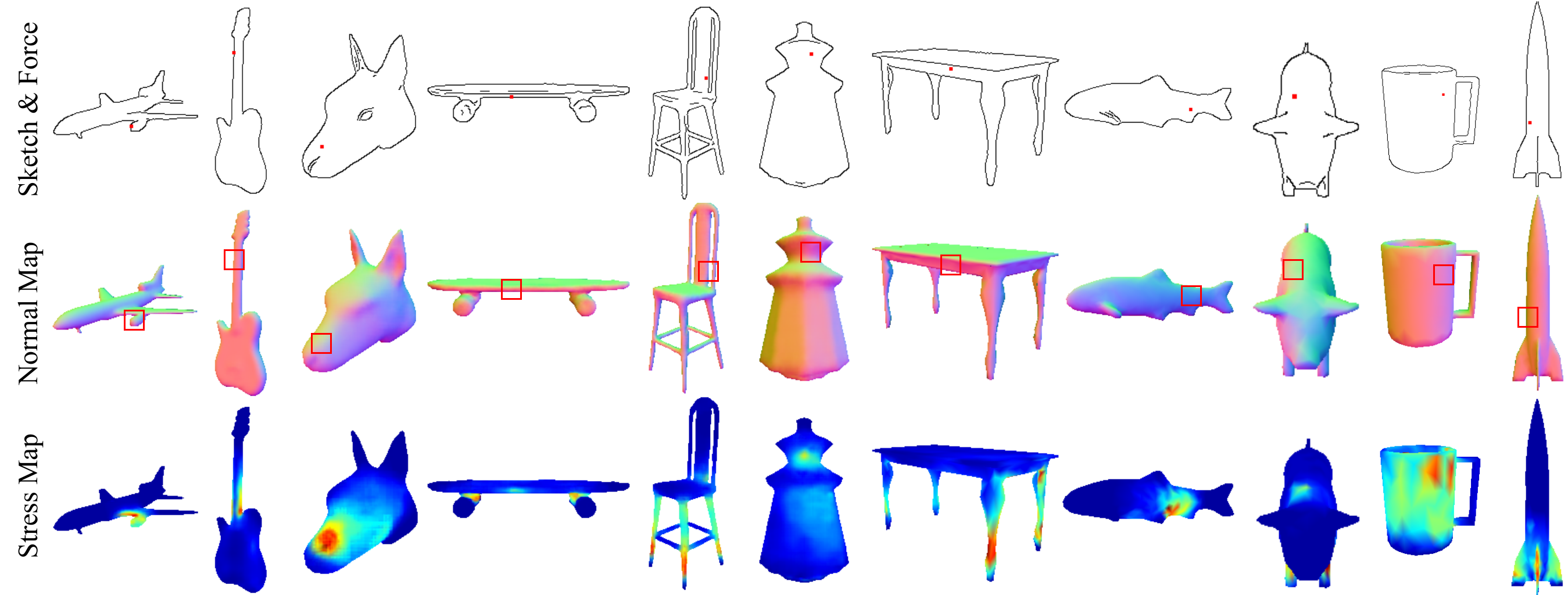}
    \caption{Result gallery of {eleven} categories in our synthetic {sketch-force-stress data}.
    The top {row shows} 
    the input sketches and external force locations (plotted as red dots), while the middle and bottom {rows} are our generated normal maps (with predicted force directions at the center of red boxes) and synthesized stress maps{, respectively}. 
    {Please zoom in to examine the details of the applied force locations{/directions} and the generated structural stress results.}}
    \label{fig:result_gallery}
    }
\end{figure*}
\subsection{Sketching Interface}
To illustrate how our proposed method aids users in analyzing and strengthening the structural weakness of their sketched objects under external forces,
we design a simple interface (Figure \ref{fig:interface}) for users to interactively edit sketches, assign {external forces at {specific} 
positions} 
to examine the stress effects, and refine their {design}. 

Our system has
two
modes (named the sketching mode and the simulation mode), 
{and they can be selected} 
through the ``Sketch" and
``Force'' {radio} buttons.
In the sketching mode, users can load their drawn sketches or directly create one {from} scratch in the ``Input" region.
We also provide several basic drawing tools for users to edit their drawings, such as clear, undo, and redo.
After finishing one complete sketch, users may change to the simulation mode.
In this mode,
users can freely impose external forces {at desired positions} 
by clicking on their drawn sketch and {examine} 
the potential weaknesses {through} 
the simulated structural stress map and the normal map.
{The auxiliary normal map { provides} 
clearer (2.5D) shape details for designers than {the input sketch and the predicted stress map} 
(most regions are in the same color, deep blue, providing limited shape details), as observed in the middle and bottom rows in Figure \ref{fig:result_gallery}.}
{During the structure refinement process,} the generated normal map can greatly help users to iteratively improve their original drawings at a fine-grained level with its provided shape details.
By iteratively using these two modes, 
designers can create their desired shapes that are also structurally sound under certain external forces.

\begin{figure*}[t]{
    \centering
    \includegraphics[width=.92\linewidth]{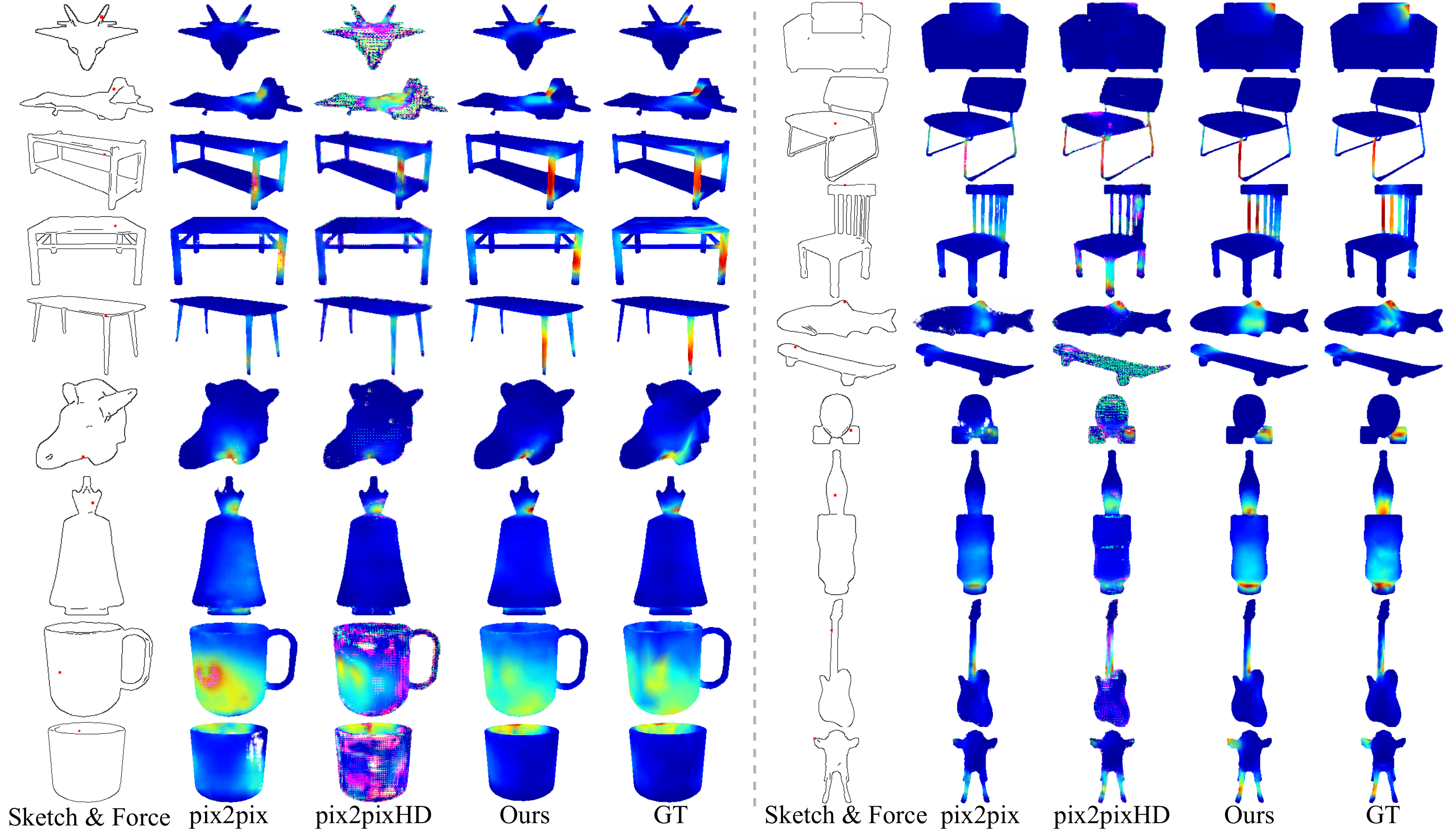}
    \caption{Qualitative comparison of {results generated by} 
    different methods of pix2pix, pix2pixHD, our method, and ground truth.
    }
   
    \label{fig:qualtative}
    }
\end{figure*}
\section{Experiments}

We evaluate our approach on {11 shape categories with a large variety of geometry and structure}, 
as shown in Figure \ref{fig:result_gallery}. 
{The 3D shapes used for sketch rendering and force-conditioned structural stress {simulation} 
are collected from the existing 3D shape repositories including ShapeNet \cite{chang2015shapenet}, COSEG dataset \cite{sidi2011unsupervised}, and AniHead dataset \cite{du2020sanihead}.}
{In total,
our synthetic dataset contains over 2.7 million sketch-force-stress data pairs with clear point-wise force annotations. The dataset spans 11 categories, namely, chairs ({1.5} million), tables (0.7 million), airplanes (0.4 million),  vases (22K), mugs (15K), skateboards (24K), rockets (4K), guitars (9K), fishes (2.6K), {four-leg animals} (13K), and {animal heads} (78K). 
After data augmentation, our collected data is able to train {our} 
neural network with satisfying generation quality. 
We provide more details of the data distribution of our sketch-to-stress dataset in Table \ref{tab:data_stress}.
}

\begin{table}[]

\begin{center}
\resizebox{\linewidth}{!}{
\begin{tabular}{l|r|r|r|r|r}
\hline
Category   & \#Shape & \#Views & \#Sketches & \#Force-points & \#Stress-map\\
\hline
Chair      & 4,277   & 3                           & 12,831                         & 1,523,390                          & 1,523,390                        \\
Table      & 3,656   & 3                           & 7,312                          & 715,566                            & 715,566                          \\
Airplane   & 2,231   & 3                           & 6,693                          & 403,926                            & 403,926                          \\
Vase       & 184     & 1                           & 184                            & 22,824                             & 22,824                           \\
Mug        & 164     & 1                           & 164                            & 15,840                             & 15,840                           \\
Skateboard & 134     & 3                           & 402                            & 24,471                             & 24,471                           \\
Rocket     & 49      & 1                           & 49                             & 4,291                              & 4,291                            \\
Guitar     & 39      & 2                           & 78                             & 9,186                              & 9,186                            \\
Fish       & 20      & 1                           & 20                             & 2,640                              & 2,640                            \\
Fourleg    & 42      & 3                           & 126                            & 13,181                            & 13,181                          \\
AniHead    & 208     & 3                           & 624                            & 78,528                             & 78,528    \\
\hline
\end{tabular}
}
\end{center}

\caption{{Data distribution of our synthesize{d} {sketch-force-stress }
dataset. The \#Shaps and \#Views refer to the number{s} of 3D shapes and projection views in different categories{, respectively}. While \#Sketches, \#Force-points, and \#Stress-map represent the number{s} of rendered 2D sketches, sampled force locations to apply external forces, and the ground-truth simulated 2D stress map{s}, respectively. }
}
\label{tab:data_stress}
\end{table}
\subsection{{Implementation Details}}
\label{subsec: implement details}
{We implemented our Sketch2Stress with the PyTorch framework \cite{paszke2019pytorch} and used the Xavier initialization \cite{glorot2010understanding}.
We {show} 
the parameter structures of the 
{two}-branch generator of Sketch2Stress in the supplemental materials. 
The entire pipeline of our Sketch2Stress was 
trained on an NVIDIA TITAN Xp GPU and optimized by the Adam  optimizer ($\beta$1 = 0.9 and $\beta$2 = 0.999) with the learning rate of $2e^{-4}$.
Here we trained our models to full convergence until the learning rate decayed to relatively small.
Note that training takes 24 $\sim$ 48 hours on a single GPU with a batch size of 16 for one category on average. 
The iteration epochs are set to $10$ for those categories with {a large number of} 
training samples, namely, Chair, Table, and Airplane.
For the {rest} 
categories, we set the training epochs to $100$, which is sufficient to achieve the satisfying generation performance in our experiments. 
Although it takes a long time to train our Sketch2Stress during the training stage due to the large size of training samples, the well-trained 
{two}-branch generator of Sketch2Stress only spends around $0.0005$ second{s} on average  to infer a structural stress map for an input sketch {under a specified force}.}

\subsection{Performance Evaluation}
Here we compare our method with {two} image-to-image baselines, i.e., pix2pix \cite{isola2017image} and pix2pixHD
\cite{wang2018high}, quantitatively and qualitatively.
We then perform {an ablation study} 
to illustrate the improvement provided by each key component in our method.
Finally, we use {three} user studies to demonstrate the practicality of our proposed method.

\subsubsection{Qualitative Evaluation}
In Figure \ref{fig:result_gallery}, we illustrate 
a number of stress maps generated from input sketches under user-specified external forces using our 
method.
It demonstrates the robustness of our method for input sketches with diverse
geometry.

We also provided visual comparisons to pix2pix and pix2pixHD  trained on our
{sketch-force-stress}
data in Figure \ref{fig:qualtative}. {In comparison with the ground truth, it can be easily seen that} 
our method achieves the best generation quality.
Among the {results generated by} 
pix2pixHD, we see obvious high-frequency noises. 
It is because the VGG loss pre-trained on the high-frequency natural images  
in pix2pixHD cannot well measure the feature difference of the low-frequency stress maps.
{Compared with the chair category, {the airplane has less training data, making the performance of pix2pixHD significantly worse.}}
pix2pix tends to lose the detail control of local regions during generation, especially surrounding the force locations. 
{More specifically, pix2pix usually fails to synthesize correct colors for the higher-stress regions but only flattens or diffuses these regions with background low-frequency colors, as shown in airplanes, chairs, animal heads, tables, and vases in Figure \ref{fig:qualtative}.
Please find more qualitative results 
in the supplemental materials.}

\begin{table}[]

\begin{center}
\resizebox{.92\linewidth}{!}{
\begin{tabular}{l|l|llll}
\hline
Category                  & Method            & MAE $\downarrow$            & EMD $\downarrow$           & FID $\downarrow$           & FM $\uparrow$            \\ \hline
\multirow{3}{*}{Chair}    & pix2pix           & 9.494         & 0.606          & 28.346         & 0.275          \\
                          & pix2pixHD         & {11.018}         & 1.316         & 75.852         & 0.186 \\ 
                          & Ours              & \textbf{9.251}         &  \textbf{0.374}         & \textbf{15.083}    & \textbf{0.412}         \\ \hline
\multirow{3}{*}{Airplane} & pix2pix           & 1.834          & 0.106          & 7.380          & 0.438          \\
                          & pix2pixHD         & 5.058         & 1.553          & 184.201        & 0.061          \\
                          & Ours              & \textbf{1.716} & \textbf{0.079} & \textbf{3.903} & \textbf{0.517} \\ \hline
\multirow{3}{*}{Table} & pix2pix           & 9.589 & 0.405         &   24.405       &  0.329        \\
                          & pix2pixHD      & 12.345 &  0.701  &  55.440     &  0.204         \\
                          & Ours           & \textbf{9.343} &  \textbf{0.321}    & \textbf{10.421}       & \textbf{0.434} \\ \hline
\multirow{3}{*}{Vase} & pix2pix           & 8.520       & 0.396         &      42.908    &   0.395       \\
                          & pix2pixHD      &   8.653      &   0.493        &  63.969       &   0.322        \\
                          & Ours           & \textbf{7.303}        &  \textbf{0.225}         & \textbf{31.647}       & \textbf{0.546}          \\ \hline
\multirow{3}{*}{Skateboard} & pix2pix           &  3.401       & 0.253         &51.883          & 0.168         \\
                          & pix2pixHD      &  6.274       &  1.731         &  315.004       & 0.044          \\
                          & Ours           &  \textbf{3.120}       & \textbf{0.091}          & \textbf{25.563}       & \textbf{0.341}          \\ \hline
\multirow{3}{*}{Rocket} & pix2pix           & 4.251        &6.011          &  72.085        &  0.029        \\
                          & pix2pixHD      &  8.168       &  4.848         &  215.600       &   0.025        \\
                          & Ours           & \textbf{3.645}        &  \textbf{0.344}         &  \textbf{54.561}      &   \textbf{0.414}        \\ \hline
\multirow{3}{*}{Guitar} & pix2pix           & 6.779        & 0.830         & 34.000         & 0.242         \\
                          & pix2pixHD      & 10.052        & 4.478          & 166.893        & 0.038          \\
                          & Ours           &   \textbf{4.983}      &   \textbf{0.188}        &  \textbf{31.481}      &  \textbf{0.455}         \\ \hline
\multirow{3}{*}{Mug} & pix2pix           &   24.116       &   1.321       &  43.275        &     0.321      \\
                          & pix2pixHD   &   25.637     &  2.362  &  112.037  &   0.205       \\
                          & Ours           &   \textbf{21.209}       &   \textbf{0.399}       &   \textbf{26.421}      &  \textbf{0.541}        \\ \hline
\multirow{3}{*}{Fourleg} & pix2pix     & 7.261     &  0.704  & 76.126  & 0.263
               \\
                          & pix2pixHD   &   7.176    &  0.664  &128.312 & 0.215 \\
                          & Ours        & \textbf{6.890}  & \textbf{0.250}  & \textbf{46.609} & \textbf{0.498}
          \\ \hline
\multirow{3}{*}{Fish} & pix2pix  &  10.988 &6.891  &124.301 &0.098
          \\
                          & pix2pixHD   & 7.094 & 1.804  &178.352 &0.123          \\
                          & Ours   & \textbf{6.209} & \textbf{0.174}  &\textbf{44.758} & \textbf{0.478}
           \\ \hline
\multirow{3}{*}{AniHead} & pix2pix  &10.399 & 0.396  &52.854 & 0.418
        \\
                          & pix2pixHD   & 9.022    & 0.650  &117.058 &0.290
   \\
                          & Ours       &   \textbf{8.535} & \textbf{0.266}  &\textbf{27.837} & \textbf{0.506}
        \\ \hline
\end{tabular}
}
\end{center}
\caption{Quantitative comparison of different methods {in {the} sketch-based structural stress generation task on the {eleven} 
categories}. 
}

\label{tab:quantitative_comparison}
\end{table}

\subsubsection{Quantitative Evaluation}
In the study of sketch-based structural analysis, we focus not only on the generation quality but also on the pixel-level {stress} accuracy of generated results{, compared with} 
the ground truth.
We adopt four metrics to comprehensively evaluate the performance of different methods and compare their generated stress maps with the corresponding ground truth, 
{namely}, 
mean absolute error (MAE), F-Measure (FM), earth mover's distance (EMD), and  Fréchet inception distance (FID).
{The former two metrics are used for the pixel-wise stress accuracy measurement, and the latter two are
for the image quality evaluation}. 
We {report} 
the quantitative evaluation {results} of 
the aforementioned methods {on {all eleven} 
categories}
in Table \ref{tab:quantitative_comparison}.
Note that we test the compared methods on the unseen data. {For instance, the test data for} 
Chair and Airplane 
contains 100 shapes and 60 shapes with $35K$ and $18K$ force samples, respectively.

\begin{figure*}[t]{
    \centering
    \includegraphics[width=.94\linewidth]{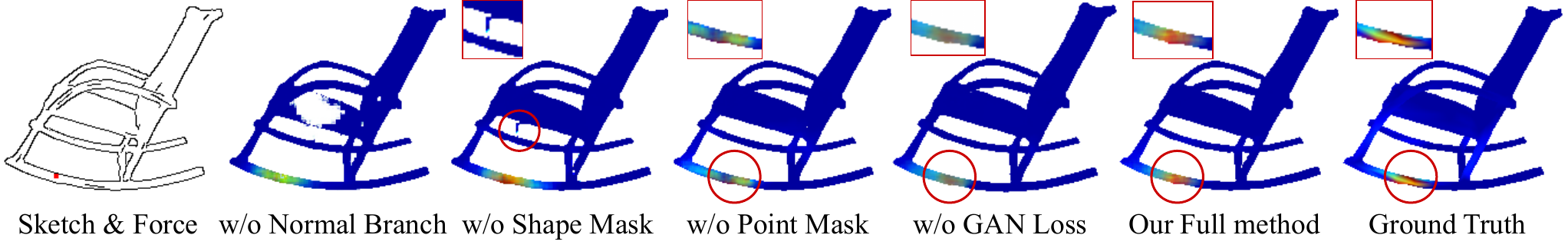}
     \caption{{Qualitative comparison of our ablated methods.}
     }
    \label{fig:ablation_study}
    }
\end{figure*}

{From} 
Table \ref{tab:quantitative_comparison}, {we observe that our method yields overall better image generation quality while achieving significant improvements in pixel-level stress accuracy compared to {the} 
competitors.
Compared to our approach,
pix2pix struggles to accurately predict stress regions surrounding the {user-specified} 
force locations, resulting in compromised image quality (FID and EMD values) and pixel-wise stress accuracy (MAE and FM values). 
For pix2pixHD, its generated stress maps contain too many high-frequency noises (see the qualitative results in Figure \ref{fig:qualtative}), leading 
to its poor performance in metrics associated with image generation quality (EMD and FID values) and pixel-wise stress accuracy (MAE and FM values).
Although the value of the four metrics fluctuates across different categories due to the different data amounts of training data, our proposed method remains consistently superior to the competitors in the sketch-based structure analysis task.}

\begin{table}[]
\begin{center}
\resizebox{.96\linewidth}{!}{
\begin{tabular}{l|l|llll}
\hline
Category                                   & Method           & MAE $\downarrow$  & EMD $\downarrow$  & FID  $\downarrow$  & FM $\uparrow$  \\
\hline
\multirow{4}{*}{Chair} & w/o Normal Branch     & 9.494         & 0.606          & 28.346         & 0.275          \\
                      & w/o Shape Mask    & 9.490    & 0.390    &    14.718 &  0.402  \\
                 & w/o Point Mask    & 9.366    & 0.378    &  15.092   & \textbf{0.415}   \\
                 & {w/o GAN Loss}    &  {9.289}   &  {0.387}   & {15.265}    & {0.392}   \\
                     & Full             & \textbf{9.251}         &  \textbf{0.374}         & \textbf{15.078}    & {0.412}        \\
\hline
\multirow{4}{*}{Airplane}                  & w/o Normal Branch          & 1.834         & 0.106          & 7.380          & 0.438     \\
                                           & w/o Shape Mask    & 2.301          & 0.126          & 4.456          & 0.457     \\
                                           & w/o Point Mask   & 1.804          & \textbf{0.078} & 4.227 & 0.514   \\
                                           & {w/o GAN Loss}    & 1.803  &  0.079   &   6.782  &  0.486  \\
                                           & Full              & \textbf{1.716} & {0.079} & \textbf{3.903} & \textbf{0.517} \\
                                           \hline

\multirow{4}{*}{Guitar}    & w/o Normal Branch  &  6.779  & 0.830  & 34.000  & 0.242    \\
                           & w/o Shape Mask   &  5.946  & 0.253  & 34.762  & 0.409    \\
                          & w/o Point Mask   & 5.178 & 0.192  & 32.982  &0.438 \\
                          & {w/o GAN Loss}    & 4.993   &  0.194   &  35.037   & 0.436   \\
                        & Full  & \textbf{4.983}      &   \textbf{0.188}        &  \textbf{31.481}      &  \textbf{0.455} \\
                                           \hline

\end{tabular}
}
\end{center}
\caption{{Quantitative comparison of the ablated methods of our approach on {three} 
categories with complex and diverse shape structures.}} 
\label{tab:ablation_study}
\end{table}

\subsubsection{Ablation Study}
{To evaluate the effects of the key components (namely, normal branch, shape mask, point mask, {and GAN loss}) of our approach,  we present both the quantitative comparison of the ablation results in Table \ref{tab:ablation_study} and the qualitative comparison in Figure \ref{fig:ablation_study}.
Note that we report the quantitative comparison of our ablated methods on the chair, airplane, and guitar categories, 
which were chosen based on their {highly} varying levels of diversity, complexity, and
{number of training structures}.
From Table \ref{tab:ablation_study}, we observe that removing the normal branch (the supervision on force directions and 2.5D shape information), as expected, has a noticeable effect on {the} four metrics, leading to 
a significant drop in image generation quality and pixel-level accuracy (also see Figure \ref{fig:ablation_study}).
Without the shape mask, our approach's performance shows a heavy decrease in the four metrics.
This mask plays a critical role in regularizing the shape boundary of the generated image{s} while also reducing the outlier noises, such as the outlier defects below the chair seat in Figure \ref{fig:ablation_study}.
In terms of the point mask, we observed a slightly poorer performance on the three categories if this component was removed. As shown in Figure \ref{fig:ablation_study}, our method tends to lose fine-grained control over regions surrounding the force point without the point mask. {After removing the GAN loss, our approach fails to learn the distribution of high-frequency pixels and produces an over-smoothed effect over high-stress regions.}

In summary, our approach relies primarily on normal maps to perceive the underlying 3D shape structures and infer the surface stress.  
The shape mask is the second most important component in supervising the shape boundaries of generated stress maps.
{Then}, the point mask plays a vital role in guaranteeing the region consistency surrounding the force points.
{Finally, the GAN loss constrains the distribution of the high-frequency pixels (high-stress regions) in the generated stress maps.}
{In our approach, the shape mask affects the stress map indirectly by regularizing the shape boundaries in normal maps directly while the point mask {and the GAN loss} influence the final stress map directly during the generation process.}
}

\begin{figure}[t]{
    \centering
    \includegraphics[width=.96\linewidth]{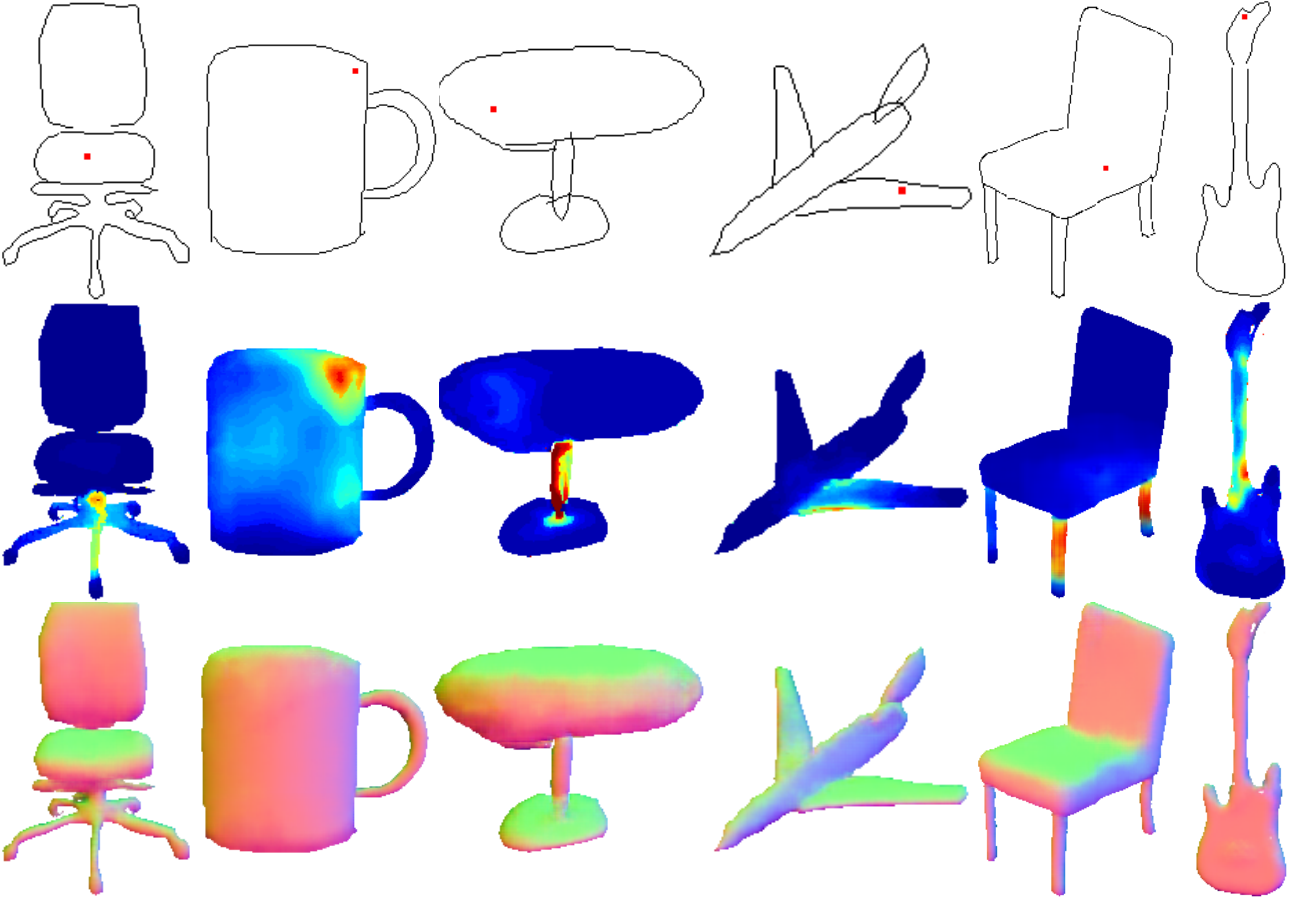}
    \caption{User study of sketch-based weakness analysis. The top row is the {user-drawn} 
     freehand sketches and the interested force points (red dots), the middle is the computed structure stress maps with our method, and the bottom is the inferred {normal maps}. {The results demonstrate that our Sketch2Stress are robust to the common defects (poorly drawn curves, imperfect straight lines, detached chair back and guitar head, and the unclosed circular table) existing in input 
     sketches and are able to generate consistent normal maps and stress effects.} 
    }
    \label{fig:user_study_structure_analysis}
    }
\end{figure}

\begin{figure*}[t]{
    \centering
    \includegraphics[width=.99\linewidth]{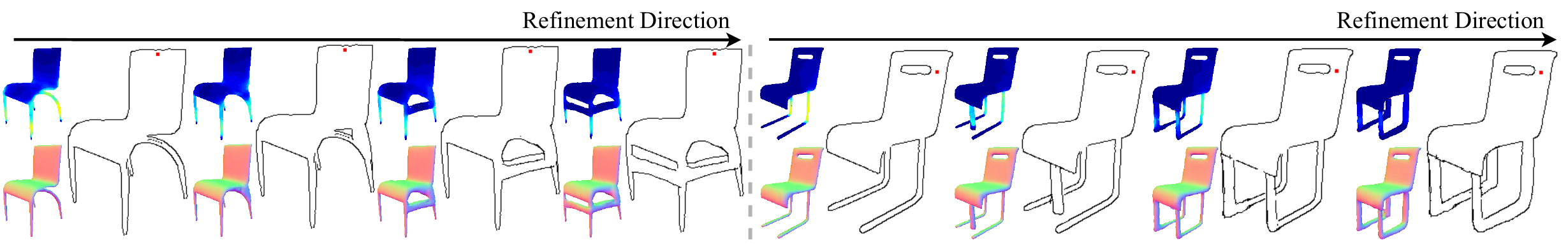}
    \caption{Two examples from the user study of sketch-based structure refinement. We display the refine{ment} directions of how users enhance the problematic structures and detail the intermediate refined sketches along the arrow{s} of {the refinement} directions.  
    The stress feedback and additional normal maps are side-placed with the sketches.
    Please zoom in to examine the details.
    }
    \label{fig:user_study_refinement}
    }
\end{figure*}
\begin{figure*}[t]{
    \centering
\includegraphics[width=.84\linewidth]{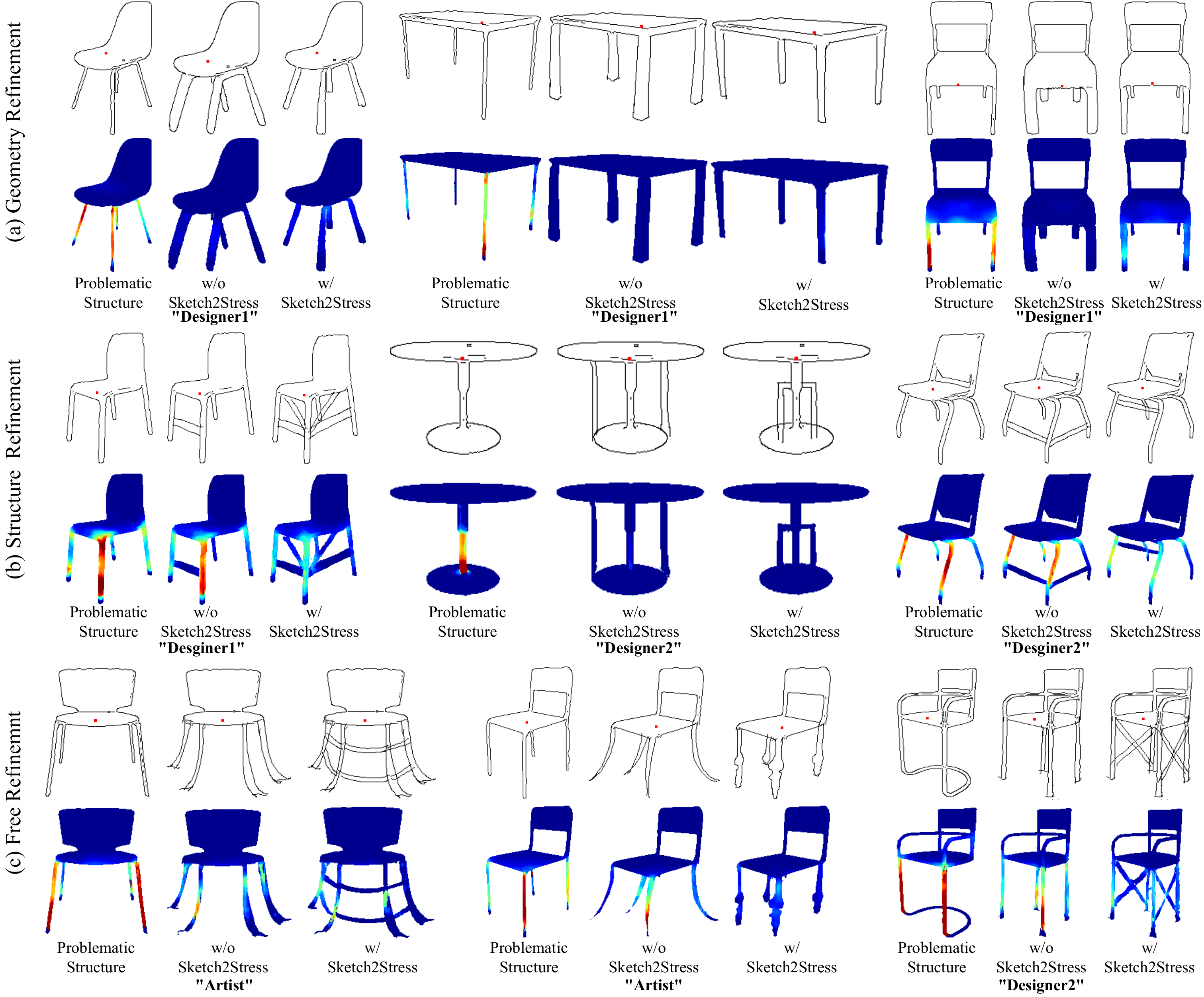}
    \caption{User study of structure refinement with or without our Sketch2Stress tool.
    Each triplet contains a {structurally} problematic 
    sketch under different force configurations (red dots on sketches), and the {user-refined results without and with our tool, respectively.} 
    The corresponding stress maps are provided under the refined sketches.  
    }
    \label{fig:user_study_with_tool}
    }
\end{figure*}

\subsection{User Studies}
\label{subsection:user_study}
To validate the practicality of our proposed sketch-based analysis tool, we design
{three} user studies. 
{The first study} is the sketch-based weakness analysis to help users to analyze and summarize 
weak regions on their freely drawn sketches.
{The second study} is the sketch-based structural refinement to help users to refine the structures of given sketches based on 
our computed stress map{s}.
{The third study examines the usefulness of our Sketch2Stress tool in assisting designers by providing them with structural stress awareness during the structure refinement process. 
Thus, we deploy a controlled {task} 
where designers are asked 
to refine the same problematic structures under the specified force conditions. 
During the trials, they are requested to refine the structure twice; first, without our Sketch2Stress tool, relying on their intuitions and experience, and second, with our tool.
We invite 9 volunteers to participate in our user studies. Two of them were professional interior designers with years of drawing experience, {and} one was {a new media} 
artist, 
and the rest were postgraduate students aged 26 to 29 with no professional drawing skills.}

\subsubsection{Sketch-based Weakness Analysis}
In this study, we invite {the participants to sketch their interested shapes among our prepared categories freely,}
with our interface, and let them figure out the fragile structural weakness by clicking on the sketched objects. 
After examining their own sketched {objects}, we require users to summarize their analysis process and answer one question ``Which kinds of regions of a shape are the potential or possible weak regions?''.
Their answers to {this question} 
are {"joint regions, thin {structures}, non-straight legs, and single legs with variable thickness"}.
We showcase  representative freehand sketches with the user-assigned forces, the corresponding stress maps, {and the inferred view-dependent 3D structures} in Figure \ref{fig:user_study_structure_analysis}. {Although the viewpoint of the 
freely sketched airplane in Figure \ref{fig:user_study_structure_analysis} is quite different from training samples (Figure \ref{fig:result_gallery}) in our dataset, our Sketch2Stress is still able to infer a faithful view-dependent structure and a feasible stress map for the input.}

\subsubsection{Sketch-based Structure Refinement}
As the chair category exhibits the most complex shape structures, in this study, we invite 
{all the participants} 
to refine and enhance {two} initial chair structures
with weak {or problematic} regions undertaking 
{higher} stress {(see the regions with warmer and lighter colors in Figure \ref{fig:user_study_refinement})} 
among the whole dataset.
During the refinement process, we do not provide any suggestions and ask users to refine the structure based on the guidance of the computed stress map in our interface.
We illustrate the refinement process of two {representative sketch-based structure refinements from users, as shown} 
in Figure \ref{fig:user_study_refinement}. 
We also display the generated stress maps and normal maps besides the refined sketches at each time step in Figure \ref{fig:user_study_refinement}.

Through the previous two studies, we show that both novice users and designers can easily identify the weak regions that sustain higher stress under the specified external forces with our proposed method.
Our method also provides an effective way for users to {interactively} enhance their created shape structures by step-by-step refinement with our generated stress maps.
However, people might be quite interested in how useful our Sketch2Stress tool is for these professional designers with years of design experience.
So we further deploy the following controlled trial{s} to answer this question.

\subsubsection{Controlled Trial{s}: Structure Refinement with or without Sketch2Stress}
{In this 
study, we invited two professional designers and one artist {among the previous study participants}  and asked them to refine problematic sketch structures as much as possible.
In the beginning, we showed the participants both a problematic sketch and its stress {map} 
which indicates the weak regions of these fragile structures (see "problematic structures" in each triplet in Figure \ref{fig:user_study_with_tool}). 

In conventional refinement tasks, designers are commonly requested to respect the original geometry (like the thickness for simplicity) and structure as faithfully as possible.
Our sketch-based structure refinement task also follows the same rule.
However, in our scenario, drastically changing the original structure by adding extra structures also works for improving the structural soundness, so we further customize and define our own requirements for different types of refinements as shown in Figure \ref{fig:user_study_with_tool}:
(a) Geometry refinement: participants are only allowed to adjust the thickness {of the fragile parts} to improve the problematic structure without {changing the original structures}. 
(b) Structure refinement: participants are only allowed to {change} 
structures but not modify the thickness of the original structures.
(c) Free refinement: participants are allowed to edit both the geometry and the entire structures.

In the first trial, guided by the requirements stated previously, the participants were asked to heal these weak regions with their own learned knowledge, design experience, and intuitions but without our Sketch2Stress. 
{Although three participants used our Sketch2Stress tool in the previous two user studies and learned where might be the potential regions and how the problematic structures were iteratively improved with our tool, we are still interested to know how well the participants could use the learned knowledge and fix the novel problematic cases by themselves. 
} 
During this process, the participants were allowed to edit the problematic sketch structures 
multiple times {following the different refinement requirements} until they were satisfied. Note that we did not update the stress maps during this refinement process.
The final refined sketches and their corresponding stress maps of the first trial can be seen in Figure \ref{fig:user_study_with_tool} ("w/o Sketch2Stress" in each triplet).
In the second trial, we allowed the participants to refine the problematic structure {obeying three refinement requirements} with our Sketch2Stress tool.
During this process, 
we provided an instant response in the form of a stress map after each editing operation. The final sketch refinements and their corresponding stress maps from the second trial can be seen in Figure \ref{fig:user_study_with_tool} ("w/ Sketch2Stress").

Through 
{the}
controlled trial{s}, 
we found that 
relying solely on designers' experience without our Sketch2Stress tool could only mildly relieve the weak regions' stress or sometimes worsen the situation. 
For example, in geometry refinement (Figure \ref{fig:user_study_with_tool} (a)), {the} designers 
usually attempted to thicken these thin legs as much as possible. However, this is not the optimal way to strengthen problematic regions meanwhile not modifying the original geometry too much.
While our Sketch2Stress tool can help {the} designers {and the artist} 
to iteratively adjust and obtain a more suitable, even optimal thickness by giving them instant stress feedback after each modification.  
Also, as illustrated in Figure \ref{fig:user_study_with_tool} (b) and (c), our  Sketch2Stress informs the designers and {the artist} where (potential fragile regions), which auxiliary strategies (thickening, adding extra structures), and how effective their modifications by showing them the instant stress responses after their edition operations.
Using Sketch2Stress, all the participants successfully refined the problematic sketch structures to better versions.
The practicality and usefulness of our Sketch2Stress tool received high appreciation from 
the designers and artist. 
All three participants {spoke} 
highly of our designed tool for helping them quickly locate the weak regions and inform them of the vivid and instant stress responses after every editing operation. {Before} 
our user studies, they did not have much experience performing structural analysis and refinement in the sketching phase.
}

\begin{figure}[t]{
    \centering
    \includegraphics[width=.98\linewidth]{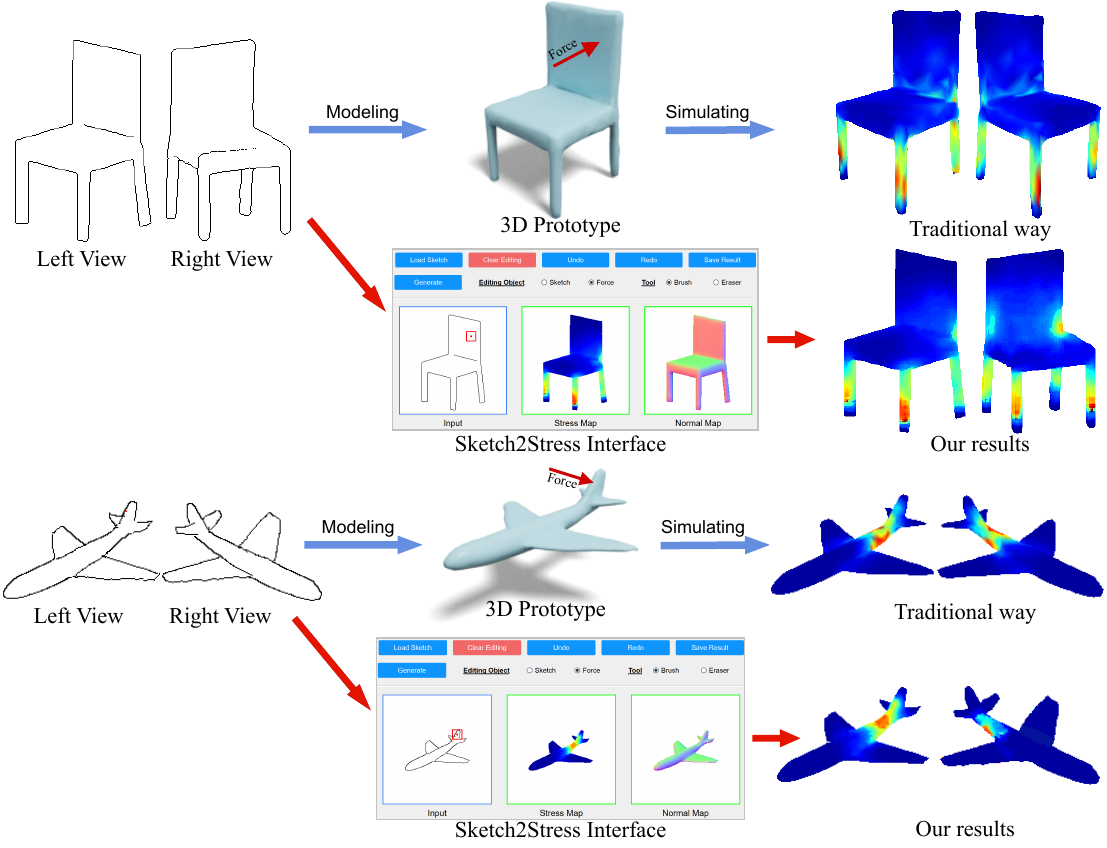}
    \caption{{The comparison of structural analysis using a traditional way or method (upper {part} {in each example}) and our Sketch2Stress approach (lower part)}.
    }
    \label{fig:user_study4}
    }
\end{figure}

\subsubsection{{Compared with 
Traditional Structural Analysis}}
\label{subsec:user_study_compare_traditional}
In {our} previous user studies, we have demonstrated the effectiveness of Sketch2Stress for novice users and professional designers in the sketch-based structural analysis and refinement tasks. In this user study, we {will further explore} 
how our proposed method {compares} with {a} traditional structural analysis {approach} {({i.e.,} the reconstruction-and-simulation approach in Figure \ref{fig:recons_and_simulate} but with multi-view input sketches)}, as illustrated in Figure \ref{fig:user_study4}. 

Following the traditional structural analysis pipeline, we first invited {two} professional interior designer{s} to draw multi-view 2D sketches of envisioned 
3D object{s} {(a chair {and an airplane} in our comparison)}, 
then {created 3D model{s} from} 
the {drawn} 2D sketches with a multi-view sketch-based reconstruction approach \cite{zhou2023ga}, and finally used the structural analysis technique \cite{ulu2017lightweight} on the resulting 3D model{s}.
Figure \ref{fig:user_study4} displays the structural stress maps produced by {such a traditional pipeline} 
and our method. 
{We also computed the distances (i.e., MAE, EMD, FID, and FM) between the resulting structural stress maps (see Figure \ref{fig:user_study4}) generated by our Sketch2Stress approach and the traditional reconstruction-and-simulation method from multi-view sketches.}
{Note that our predicted stress maps for the multi-view input sketches are generated separately since our method is designed for the single-view scenario. Therefore, the generated stress maps by our method may lack
the consistency across views (see the inconsistent stress effects across views in regions of the chair back and the airplane tail in Figure \ref{fig:user_study4}), compared to the traditional way.}
{From this figure,} 
we can observe that given the 
{individual left-view} {sketches} {of both the chair and the airplane}, our result{s} 
generally reflect 
{the} consistent stress distribution comparable to the result{s} by 
the traditional method, {though} 
there are some geometrical distortions on the 
{chair legs and back, and the airplane tails and wings}
in the generated result{s} due to the imprecise sketching.
In addition, the {view angle of the} right input view{s} {(drawn at around $-45$ degrees azimuth angle) is a novel view that was not 
in our dataset ([0, 45, 90] degrees in Section \ref{subsection:data_rendering}) and 
never }
seen by our model
{(see the changes 
in the computed distances when switching from the left to the right views in Table \ref{tab:comparison_sk2stress_and_traditional_way})}.
However, our method can still infer 
similar and reasonable stress map{s} compared to the traditional method {(see Figure \ref{fig:user_study4})}. 
{From Table \ref{tab:comparison_sk2stress_and_traditional_way}, we also observe that the distances between our method and the traditional method change significantly with input of sketches with different levels of distortions 
(the fewer distortions (airplane sketch), the smaller distances). This indicates the importance of professional drawing skills and the necessity of further beautification for the user's freehand sketches in the early sketching stage of the design and fabrication processes \cite{yu2023sketch}.}

\begin{table}[]
\begin{center}
\resizebox{.94\linewidth}{!}{
\begin{tabular}{l|l|llll}
\hline
Examples  & Views      & MAE $\downarrow$  & EMD $\downarrow$  & FID  $\downarrow$  & FM $\uparrow$  \\
\hline
\multirow{2}{*}{Airplane} & Left View  & \textbf{3.842}    & \textbf{0.166}    & \textbf{49.074} & \textbf{0.499}         \\
                          & Right View    & 4.123    & 0.348    &  125.668   &  0.357  \\
\hline
\multirow{2}{*}{Chair}    & Left View     & \textbf{18.074}   & \textbf{1.072}    & \textbf{192.250} & \textbf{0.404}    \\
                          & Right View    & 29.765    & 1.401    & 299.330   & 0.178  \\
\hline
\end{tabular}
}
\end{center}
\caption{{Quantitative evaluation of the distances between our proposed approach and the traditional analysis method with input with seen (left) and unseen (right) views and freehand sketches with different levels of distortions (see the freely sketched chair and airplane in Figure \ref{fig:user_study4}).}} 
\label{tab:comparison_sk2stress_and_traditional_way}
\end{table}
The similar results between the generated stress maps by our method and the traditional 
method confirm the effectiveness of our 
approach.
{Furthermore, in the traditional way, although the quality of the generated mesh and simulated structural stress (Figure \ref{fig:user_study4}) with multi-view 
reconstruction 
is significantly better and more faithful to the input sketches than 
single-view 
reconstruction 
(Figures \ref{fig:sk2mesh} and \ref{fig:recons_and_simulate}), in practice, it is still mentally demanding for 
designers {to accurately depict a desired 3D shape with multi-view sketches}.
{In contrast, our method} directly infers the corresponding stress map from an input sketch (far less than 1 second, as quantified in Section \ref{subsec: implement details}), {thus making designers bypass}
the cumbersome reconstruction and simulation steps (at least 10 seconds for each user-assigned force, even with our fastest deployment). 
{With our method, designers}  can rapidly obtain a sufficiently good structural stress map of input sketches{, without considering}
the complex spatial relationships among multi-view sketches}. 
{Through the comparison between our method and the traditional way,} we further show that it is feasible for designers to perform the structural analysis task in the early sketching stage.


% ...
\subsection{{Structural Analysis on Real Product Sketches}}

\begin{figure}[t]{
    \centering
    \includegraphics[width=.98\linewidth]{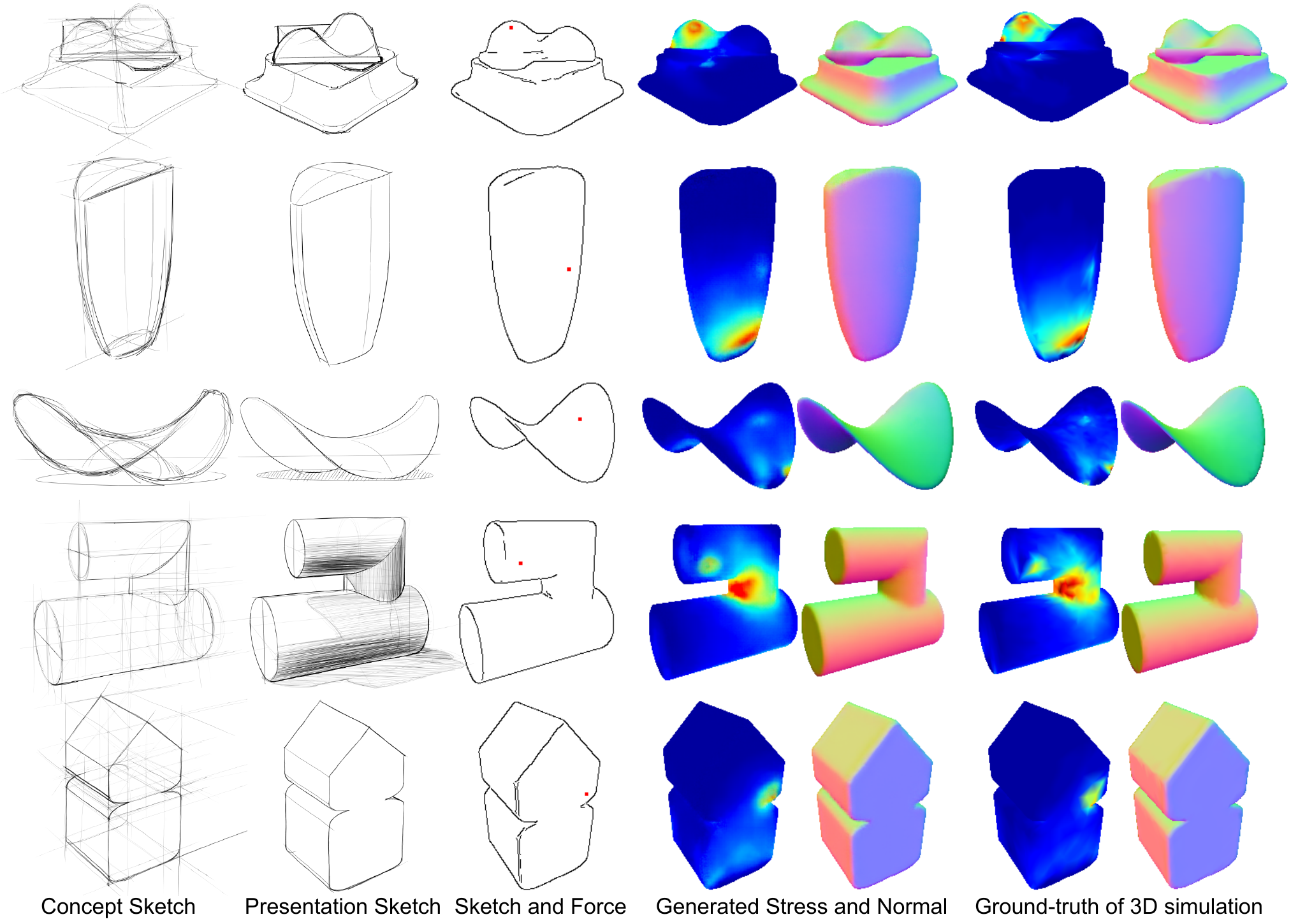}
    \caption{Our Sketch2Stress method applied to the OpenSketch dataset. The  concept and {presentation} 
    sketches of the bump, shampoo bottle, and potato chip (in the first, second, and third row{s}) are from "Professional1" while the bottom 
    {two rows}
    of the tube {and the house} are from "Professional5" {and "Professional6"} in the OpenSketch dataset.
    Please zoom in to examine the details.}
    \label{fig:opensketch}
    }
\end{figure}

As product designers extensively use sketches in their creation and communication, 
to demonstrate the powerful feature of our method in aiding sketch-based structural analysis,  
we further
apply our Sketch2Stress method to real product design sketches in OpenSketch \cite{gryaditskaya2019opensketch}.

{More specifically, w}e first leverage the 3D objects in OpenSketch to render the 2D 
{sketch-force-stress} 
data 
(as {described} 
in Subsection \ref{subsection:data_rendering}) and then train our network on the projected synthetic data. 
{The OpenSketch dataset has a
limited {number} 
of shapes {with} 
highly diverse 
structures. To conduct the shape-level 
training and testing {(in contrast to the category-level training and testing in Table \ref{tab:data_stress})}, for each shape, we rendered 12 views instead of 3 views of the previous 11 categories.
We uniformly sampled 70\% of the force points from all the force points and took their corresponding stress maps in each view as the training set, {and used the remaining force points for testing our trained model}. 
This training strategy is implemented based on our Observation (ii) {(Section \ref{section:intro})}{, i.e.,} 
the neighboring force points tend to produce similar structural stress responses.}

As shown in Figure \ref{fig:opensketch},
with clean sketches and the user-assigned forces, 
our {re-trained model}
is able to generate 
feasible and high-quality structural stress map{s} for {the forces applied on the real} product sketches.
With the aid of our method, designers will have more opportunities 
to check and refine the structural weaknesses of their ideal products in advance at the sketching stage. 
Furthermore, with our method, designers will have a larger design space by incorporating external physical factors in the form of different force configurations.






% ...


% ...


% ...

\section{Conclusion and Discussion}

We have introduced the novel problem of sketch-based structural analysis, where we constrain the external forces to variables with the same magnitude but different locations and opposite-normal directions.
We further present a 
{two}-branch 
generator to synthesize feasible structural stress maps by considering the sketches' geometry and force variables simultaneously.
We find that usually, the long, thin, tilt, and joint regions tend to suffer higher stress, and shapes with such regions are weaker than those shapes without {them}.

\begin{figure}[t]{
    \centering
    \includegraphics[width=.84\linewidth]{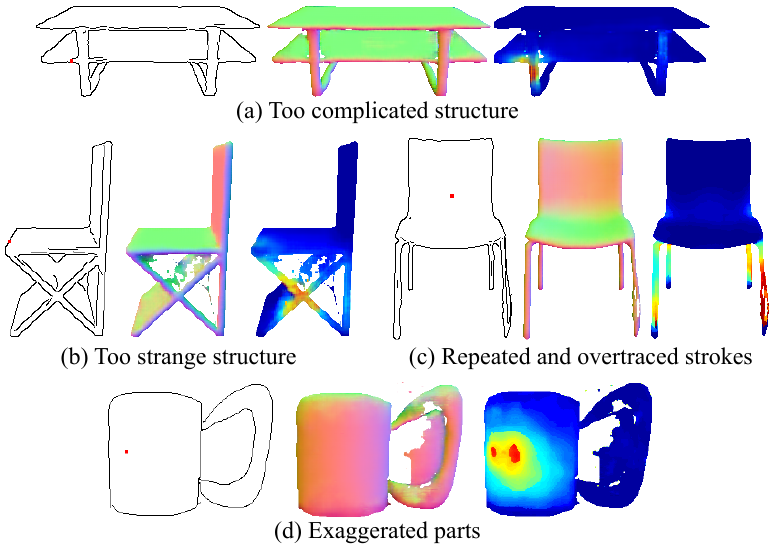}
    \caption{{{Failure} 
    cases. Each triplet contains the input sketch with the specified force {point}, 
    the generated normal map,
    and the inferred stress map. Our inference model might fail when the sketched structures are too complex (a), strange {(b), containing over-traced strokes (c), or with exaggerated part geometries (d)} 
    compared with the observed 
    {samples}
    in our dataset.}
    }
    \label{fig:limitation}
    }
\end{figure}

While the work we proposed provides an efficient approach for sketch-based structural analysis, our method 
has 
some limitations. 
{First}, our method cannot synthesize the stress effects of forces that are not in opposite-normal directions.
{Second}, the force magnitude in our problem is set to a fixed value, which makes it challenging to analyze the stress effects of external forces with dynamic values.
These two limitations are inherited from the method \cite{ulu2017lightweight} we adopted for synthesizing the training data. Hence, further advances in new structural stress analysis solutions on 3D models{, such as more efficient structural analysis techniques\cite{umetani2013cross}}, could also help to improve our approach.
{Third}, since our Sketch2stress method is learning-based,
it might fail to infer reasonable stress maps and faithful underlying structures {for input sketches} 
{with} {strange
structures, repeated and over-traced strokes, or exaggerated part geometries.} 
As shown in Figure \ref{fig:limitation} (a) and (b),
some defects can be observed in the generated normal maps and synthesized structural stress maps of the double-layer table and "X"-leg chair.
Also, the over-traced strokes and the exaggerated parts will lead to failures with our approach, i.e., the holes and the mismatched mug-handle in (Figure \ref{fig:limitation} (c) and (d)). 
For the small flaws in generated stress maps and normal maps in Figure \ref{fig:limitation} (c) and (d), they could be fixed by refining the normal map (similar to \cite{su2018interactive}).
{Lastly, similar to \cite{zhang2021sketch2model, guillard2021sketch2mesh, mescheder2019occupancy}, we train individual models for each category {to exploit} 
the finer-grained structural similarities and subtle structural differences in the same category. 
Therefore, when deploying our approach, the weights of different categories have to be reloaded accordingly during the inference time.
However, in practice, a universal network trained to handle diverse kinds of objects is more welcomed{, though it might be at the cost of a longer training time and decreased performance}.
{In particular, for freeform sketching and ideation at the early stage in product design and digital fabrication, a universal network is more friendly for designers to have more creative space and explore 
more fancy interactions of objects from 
different classes. 
}}

\subsection{{Future Work}}
{In the future, we would like to improve our 
approach from the following two aspects, i.e., practical usage and more accurate synthesis.

For practical usage, in our user studies, users needed to draw their refinements on our 
interface repeatedly by
trial and error to obtain {refined} 
structures.
Ideally, a more intuitive interface
will further guide users to fix the problematic parts {more easily}, 
e.g., by providing a slider for users to adjust the thickness {of problematic parts}.
Furthermore, to simplify the setup of the stress {map} computation {or} 
approximation, our current method re-orients the sample objects in their upright positions.
In practical applications, users might need to re-orient the objects in the desired
directions
and perform re-training.
In the future, we would like to extend our 
approach to allow for more user control over orientations.
Our 
method is {currently} designed in the single-view scenario{, which} 
cannot guarantee the 
consistency across views in
the predicted stress maps (Section \ref{subsec:user_study_compare_traditional}). 
{We might}
combine our approach with sketch correspondence algorithms (e.g., SketchDesc \cite{yu2020sketchdesc}) to further compute the dense correspondences among multi-view sketches to reduce such inconsistency.}
Only having one sketch provides limited information to indicate the material properties.
{Meanwhile}, our {current} approach cannot take as input multi-forces at different directions since the combination of multiple forces requires an extra module to process carefully, not simply recording the mapping between input multiple forces and the {corresponding} output 
structural stress effect.
In the future, we aim to explore further the proper representations and definitions for materials and external {multiple} forces in the
generative process.

{For more accurate synthesis, compared to the traditional structural analysis method (Section \ref{subsec:user_study_compare_traditional}), our method can produce coarse-level comparable structural stress results 
and be {potentially} used as an upstream process for 3D fabrication{, e.g., structural enhancement \cite{stava2012stress,miki2015parametric}.}
For 
more fine-grained level fabrication, it is still challenging for our method to generate 
precise stress distributions, especially the orientations of the stress in a generated stress map.
This can 
be {partially} solved by incorporating mechanisms from 
more advanced frameworks for high-quality image generation, e.g., diffusion models \cite{sohl2015deep,ho2020denoising,song2020score}.
Furthermore, in our current method, only the stress magnitude (low-order) of the sketched structure {but not the stress tensor (second-order)} is considered. 
Therefore, 
exploring {a} 
proper representation for the stress tensor of a sketched structure is also a promising direction to further boost 
{the accuracy of synthesized stress maps}. }

\section*{Acknowledgements} 
We thank the anonymous reviewers for their constructive comments. This work was partially supported by grants from the Research Grants Council of the Hong Kong Special Administrative Region, China (No. CityU 11212119, 11206319, and 11205420) and the Centre for Applied Computing and Interactive Media (ACIM) of the School of Creative Media, CityU.

\ifCLASSOPTIONcaptionsoff
  \newpage
\fi



\bibliographystyle{IEEEtran}
\bibliography{sections/skstressbib}
%
  

%

\begin{IEEEbiography}[{\includegraphics[width=1in,height=1.25in,clip,keepaspectratio]{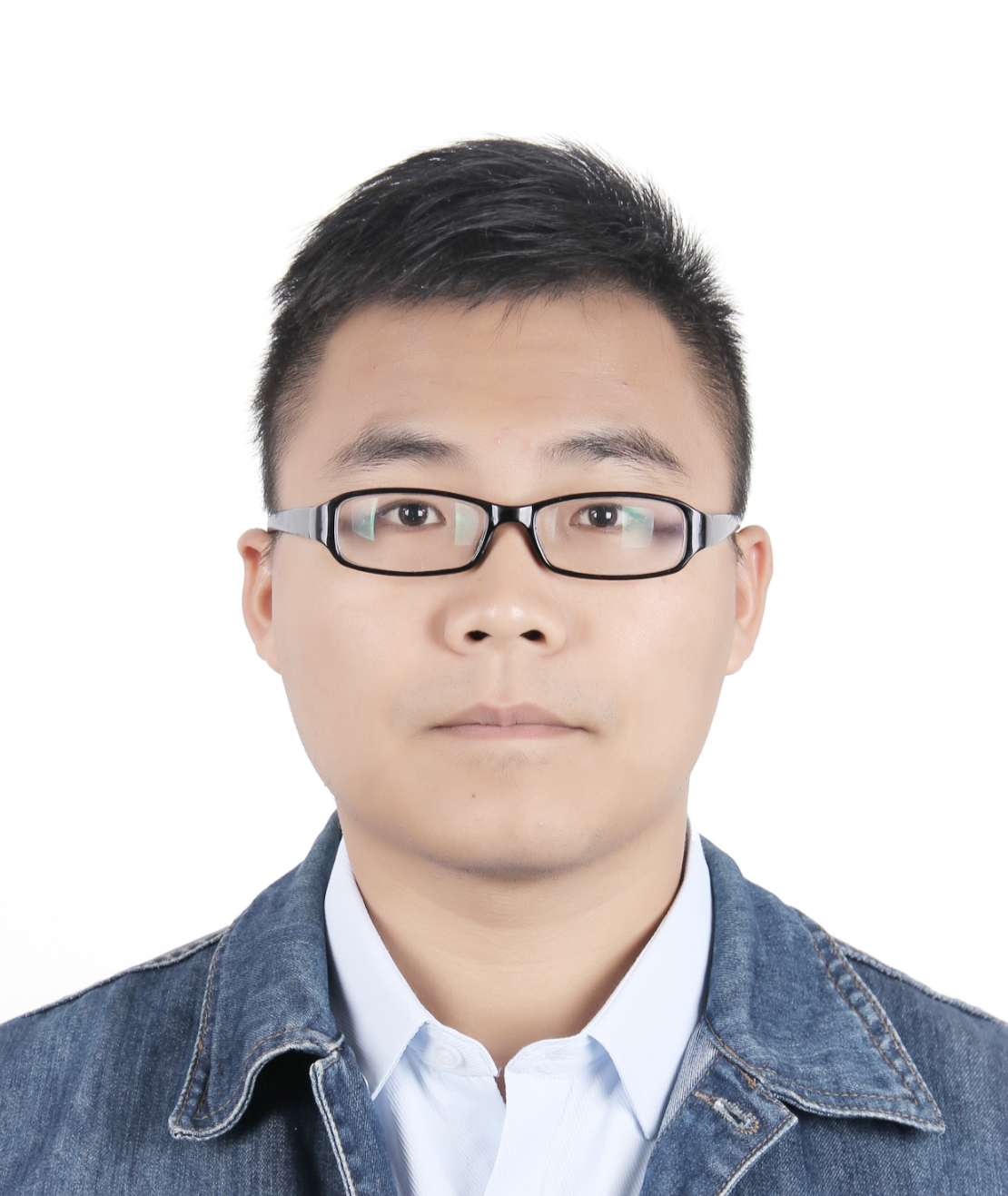}}]{Deng Yu}
is pursuing the Ph.D. degree at
the School of Creative Media, City University of Hong Kong. He received the B.Eng. degree and the Master degree in computer science and technology from China University of Petroleum (East China). His research interests include computer graphics and data-driven techniques.
\end{IEEEbiography}
\begin{IEEEbiography}[{\includegraphics[width=1in,height=1.25in,clip,keepaspectratio]{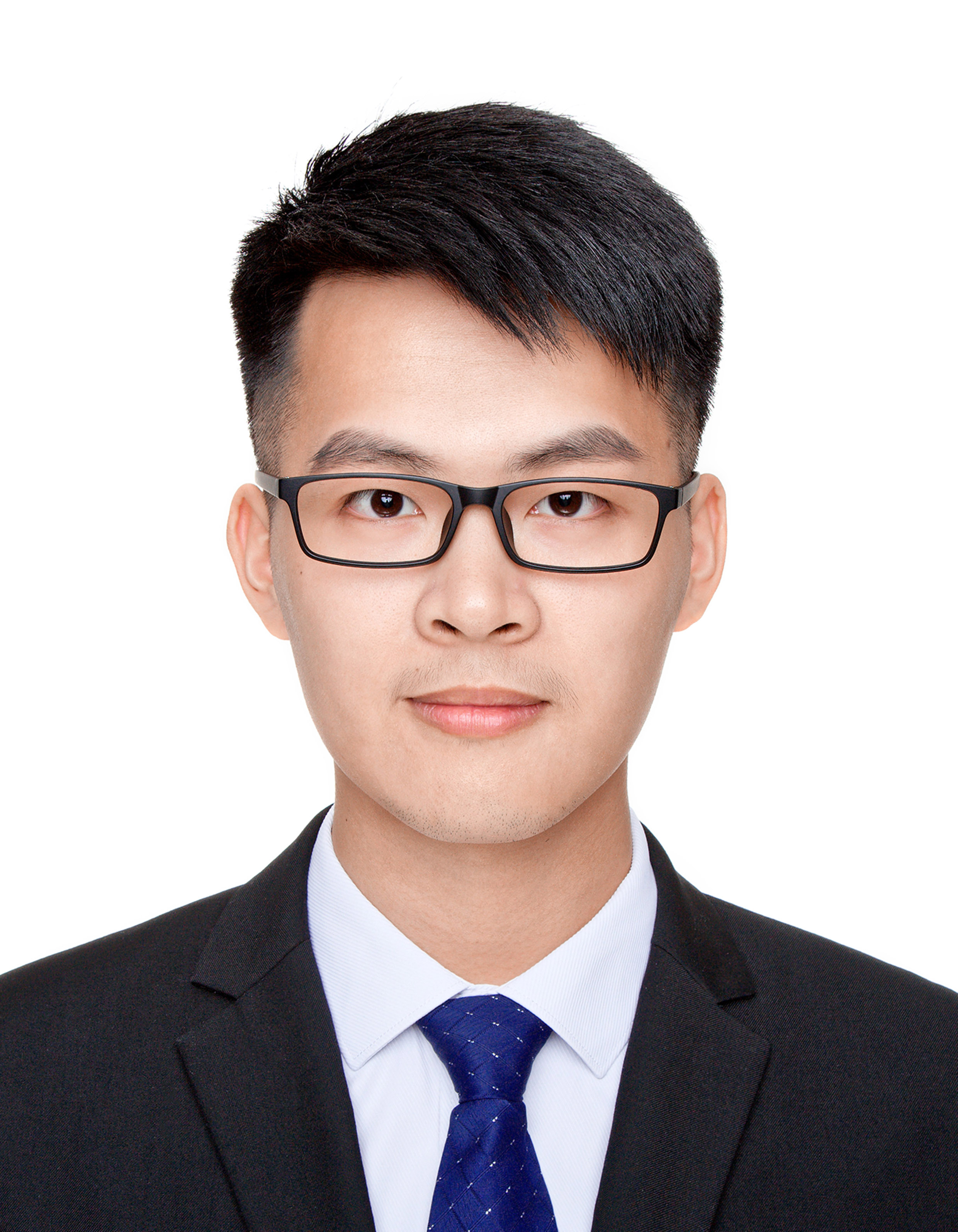}}]{Chufeng Xiao} is working toward the Ph.D. degree at the School of Creative Media, City University of Hong Kong. Before that, he received B.Eng. degree in computer science and network engineering from South China University of Technology. His research interests include computer graphics and human computer interaction.
\end{IEEEbiography}
\begin{IEEEbiography}[{\includegraphics[width=1in,height=1.25in,clip,keepaspectratio]{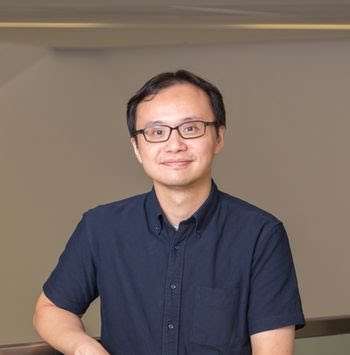}}]{Manfred Lau}
is an Assistant Professor in the School of Creative Media at the City University of Hong Kong. His research interests are in computer graphics, human-computer interaction, and digital fabrication. His recent research in the perception of 3D shapes uses crowdsourcing and learning methods for studying human perceptual notions of 3D shapes. He was previously an Assistant Professor in the School of Computing and Communications at Lancaster University in the UK, and a post-doc researcher in Tokyo at the Japan Science and Technology Agency - Igarashi Design Interface Project. He received his Ph.D. degree in Computer Science from Carnegie Mellon University, and his B.Sc. degree in Computer Science from Yale University. He has served in the program committees of the major graphics conferences including Siggraph Asia.
\end{IEEEbiography}
\begin{IEEEbiography}[{\includegraphics[width=1in,height=1.25in,clip,keepaspectratio]{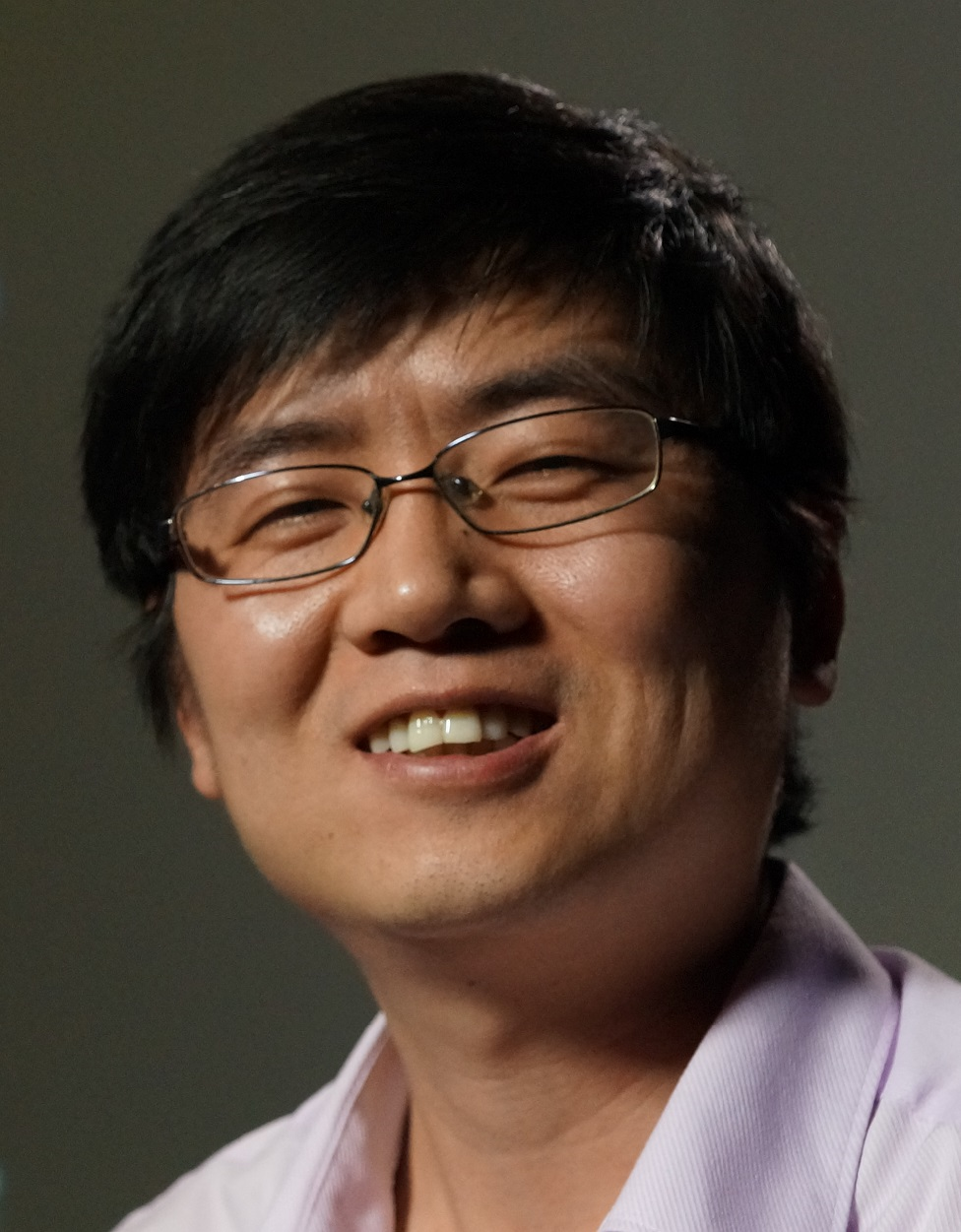}}]{Hongbo Fu}
is a Professor with the School of
Creative Media, City University of Hong Kong.
He received the B.S. degree in information sciences from Peking University, and the Ph.D.
degree in computer science from Hong Kong
University of Science and Technology. He has
served as an Associate Editor of The Visual
Computer, Computers \& Graphics, and Computer
Graphics Forum. His primary research interests
include computer graphics and human
computer interaction.
\end{IEEEbiography}
\vfill
\vfill
\vfill
\vfill
\vfill
\vfill

\vfill



\section*{Supplementary material}

# Supplemental Materials
# Sketch2Stress: Sketching with Structural Stress Awareness

| Category | Method | MAE ↓ | EMD ↓ | FID ↓ | FM ↑ |
|---|---|---|---|---|---|
| | Branchless Decoder | 9.848 | 1.319 | 33.427 | 0.195 |
| Chair | Coordinates Input | 10.892 | 0.386 | **14.534** | 0.395 |
| | Ours | **9.251** | **0.374** | 15.078 | **0.412** |

TABLE 1
Quantitative comparison of the variations of our approach.

## 1 METHOD

We adopt a framework of a two-branch generator and multi-scale discriminators to synthesize the structural stress map directly from the inputs of sketches and the user's specified force configurations. The two-branch generators are designed as an encoder-decoder architecture, as shown in Figure 1. For multi-scale discriminators, we adopted the same pixel discriminators as pix2pixHD to distinguish the real/fake multi-scale normal maps and stress maps.

## 2 EXPERIMENTS

### 2.1 Two-branch Decoders VS. Branchless Decoder

We compare our two branch decoders outputting the normal and stress maps, respectively, with the branchless decoder producing a single output concatenating multi-channels of the stress map, the normal map, and the mask image. As illustrated in Figure 2, there are usually high-frequency noises in the generated stress map by the branchless decoder. In our view, this is due to the incapability of the discriminator that can not easily distinguish the multi-channel concatenated output as fake or real. With two branch decoders and two discriminators for stress maps and normal maps, our network structure introduces more parameters to learn the force-sketch-stress mapping. It produces clearer normal maps and stress maps at a fine-grained level compared with the branchless decoder. Table 1 further displays the quantitative comparison of the two methods.

### 2.2 Point Map Input VS. Two-digit Coordinates input

To demonstrate the effectiveness of the input point map and its computed point mask in representing a user-assigned force location, we compare the synthesized stress maps of the chair category based on the aforementioned input with those directly based on the two-digit coordinate of the force location. We trained a network with the same two-branch

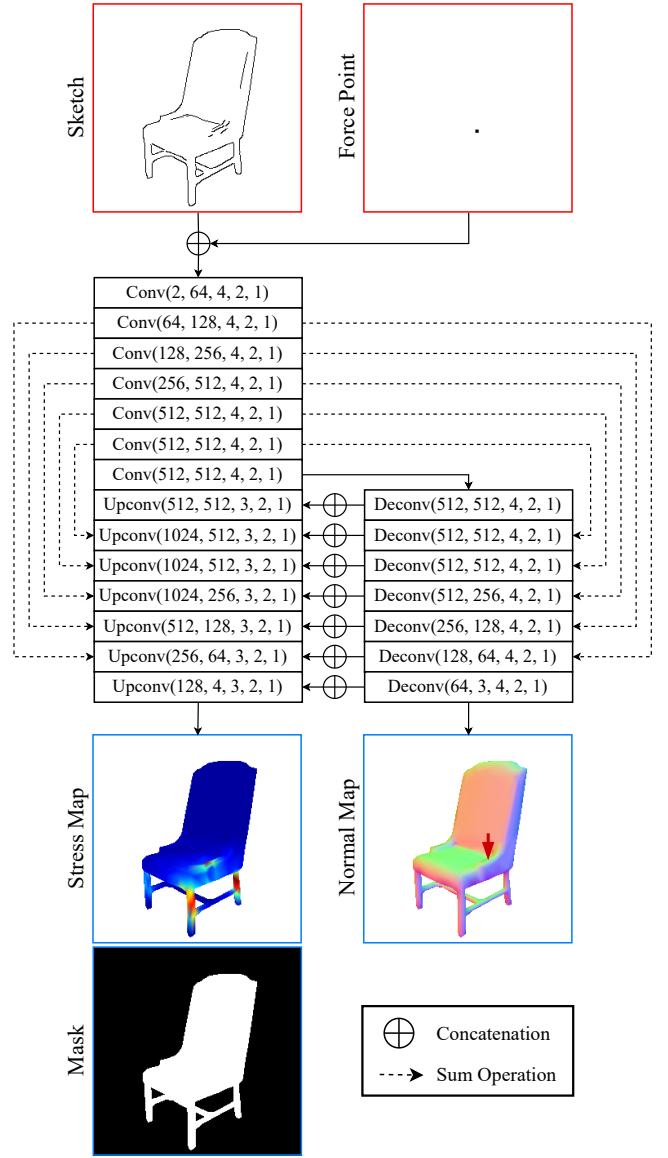

Fig. 1. The details of our two-branch Sketch2Stress network. The block of *Conv/Upconv/Deconv (input_channel_number, output_channel_number, kernel_size, stride, padding_width)* respectively represents Convolution, Upsampling + Convolution, and Deconvolution layer. Note that each Conv/Upconv/Deconv layer is followed by leaky ReLU and batch normalization, which are omitted in the figure for simplicity.

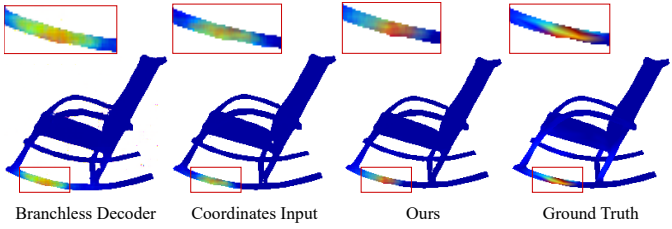

Fig. 2. Compared with some variants of our proposed method. Please see the sketch and specified force point in Figure 11 of the main paper.

and universality of our method, we give examples where the force point is applied at different object parts and further away from the weak regions. As shown in Figure 3, our method still performs significantly better than the compared two-digit coordinates input method. This indicates that our method can identify the force point more clearly than the compared method with the help of the designed point mask.

### 2.3 More Results

To demonstrate the performance of all competitors in the sketch-based structural stress generation task, we display more qualitative results in Figure 4, Figure 5, and Figure 6. Note that all the methods are tested on the unseen data.

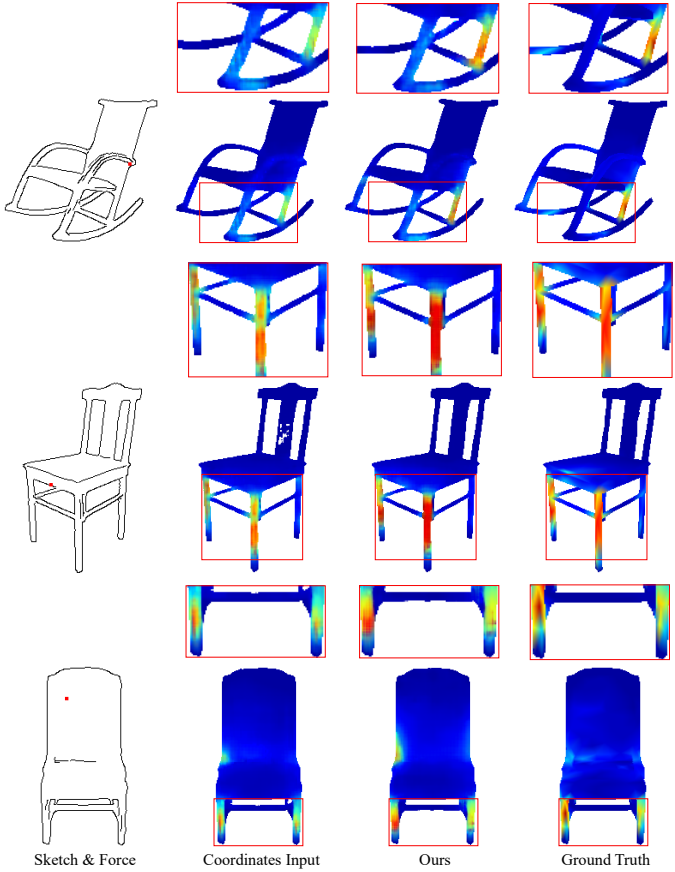

Fig. 3. Comparison of weak regions that are far from the assigned force point (red dot on sketches).

architecture while taking as input a sketch ($256 \times 256$) and coordinates of the force location($1 \times 2$) in the $256 \times 256$ space. Then, we fed the sketch to the encoder to extract its feature and concatenated the sketch feature and the two-digit force location. Finally, we decoded the concatenated feature through the normal and stress branches to produce the final normal and stress maps. Figure 2 shows the computed structural stress maps of methods with the coordinates input and point map input. Compared to our method, we observed that only providing the two-digit coordinates to indicate the force location tended to lose fine-grained control over the regions surrounding the force location in the predicted stress map. Table 1 also illustrates the quantitative performance of the two methods.

The weak regions and the force point are close in the example in Figure 2. To further illustrate the effectiveness

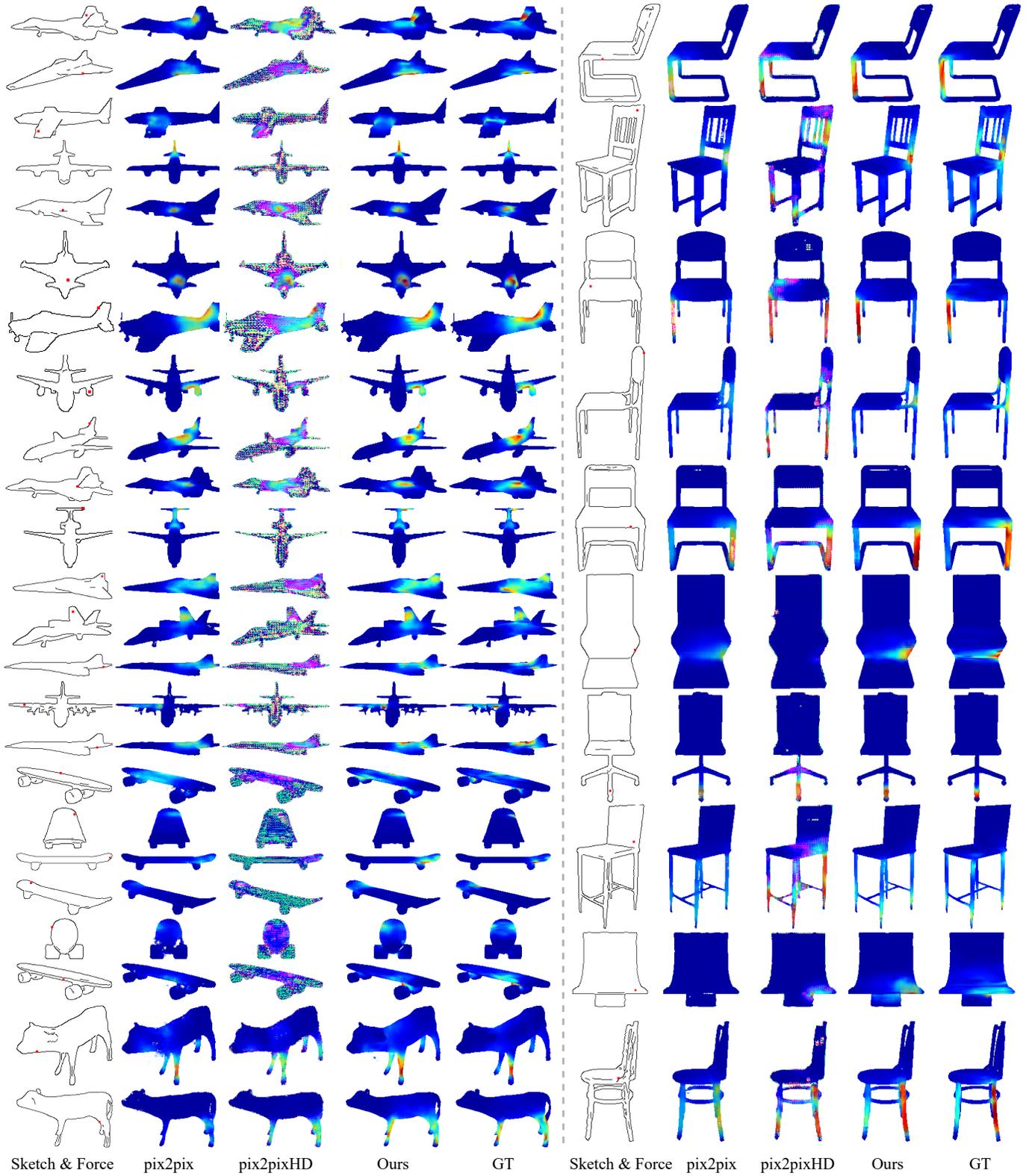

Fig. 4. Qualitative comparison of results generated by different methods of pix2pix, pix2pixHD, our method, and ground truth. The leftmost column shows the input sketches and the external forces (plotted as red dots on sketches).

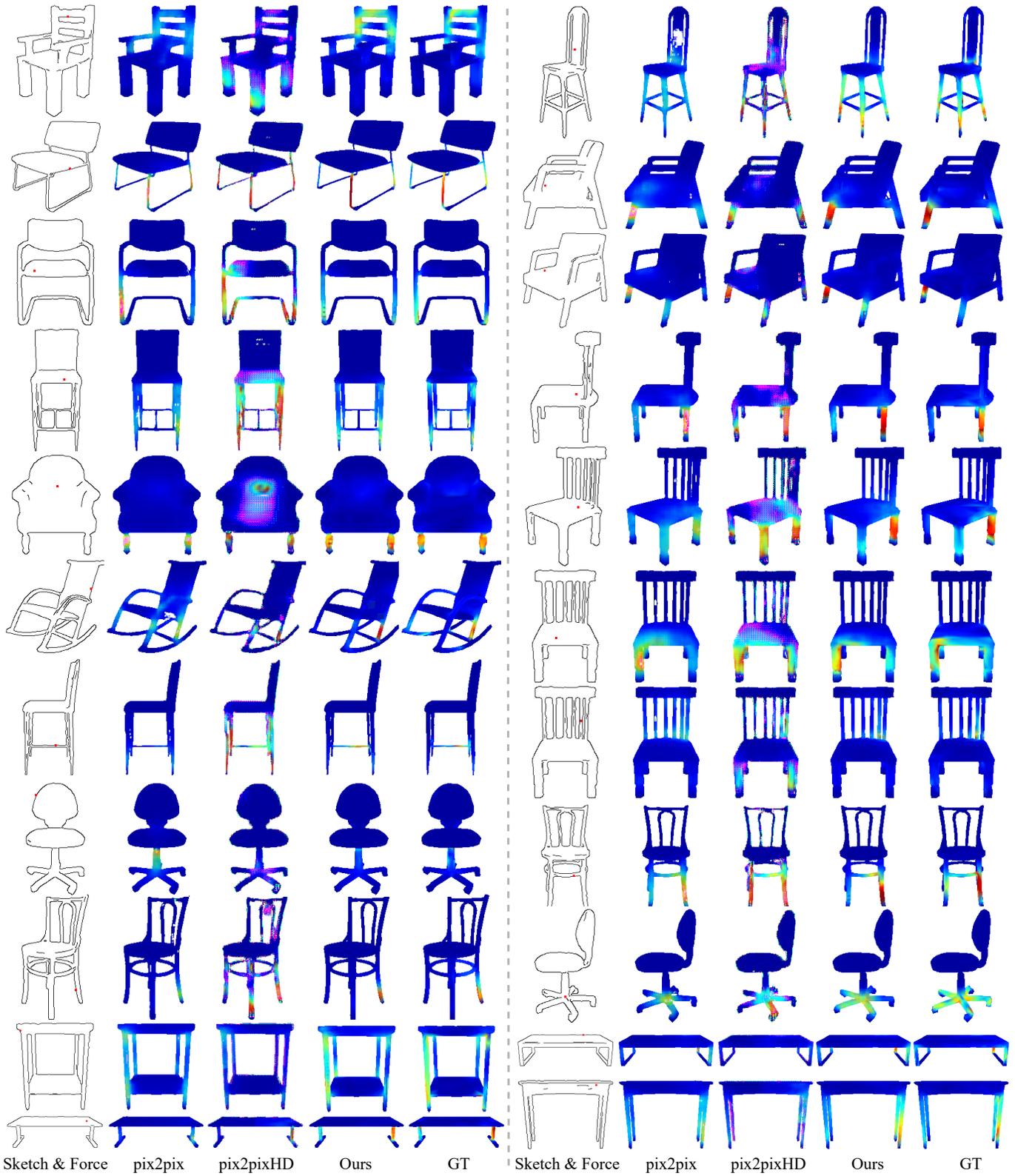

Fig. 5. Qualitative comparison of results generated by different methods of pix2pix, pix2pixHD, our method, and ground truth. The leftmost column shows the input sketches and the external forces (plotted as red dots on sketches).

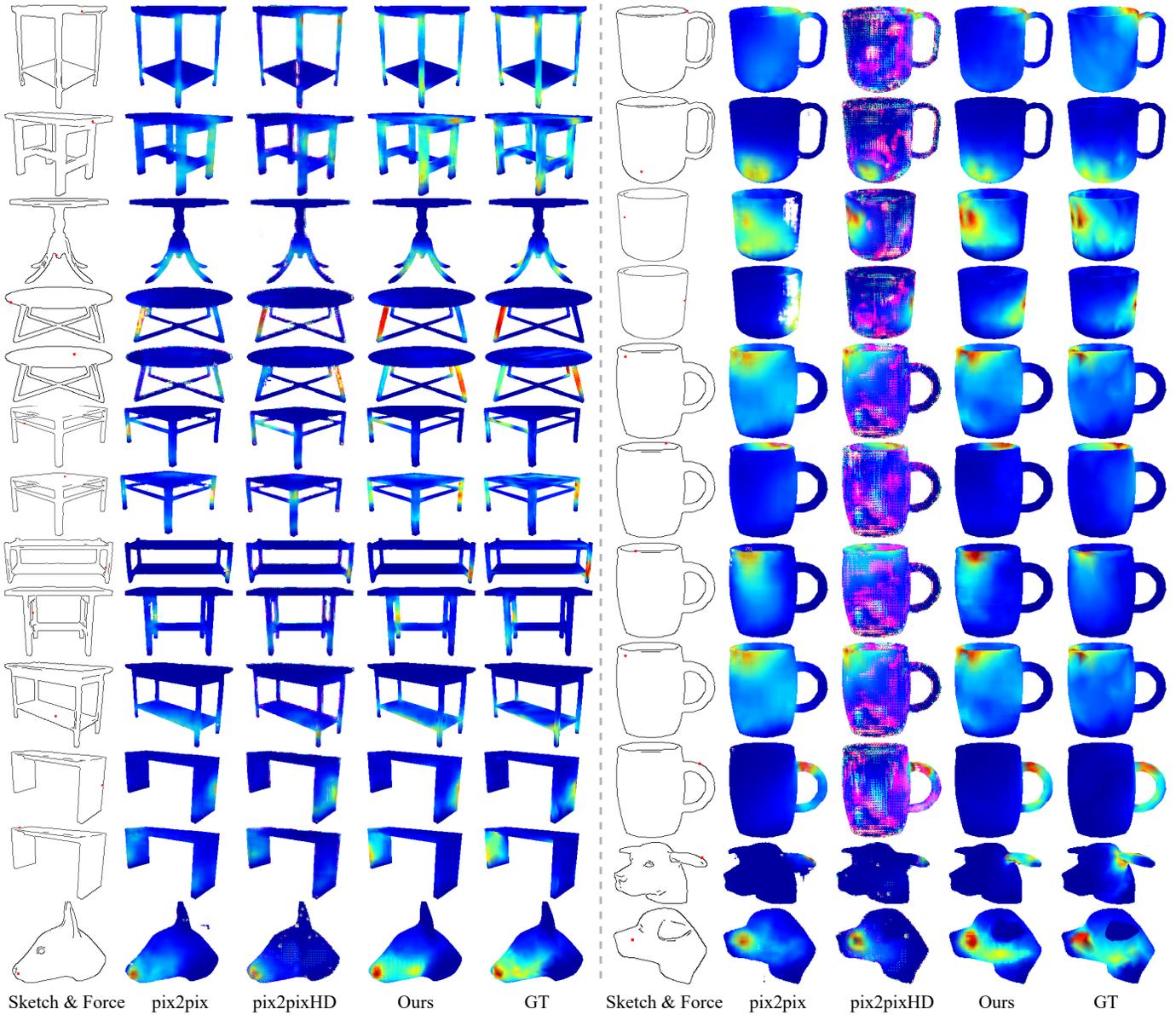

Fig. 6. Qualitative comparison of results generated by different methods of pix2pix, pix2pixHD, our method, and ground truth. The leftmost column shows the input sketches and the external forces (plotted as red dots on sketches).



\end{document}